%% file: main.tex
\documentclass{article}

\PassOptionsToPackage{square,numbers,compress}{natbib}
\usepackage[main,final]{neurips_2026}

\usepackage[utf8]{inputenc}
\usepackage[T1]{fontenc}
\usepackage{url}
\usepackage{booktabs}
\usepackage{amsfonts}
\usepackage{nicefrac}
\usepackage{microtype}
\usepackage{needspace}
\usepackage{xcolor}

\input{preamble}

\usepackage[breaklinks,colorlinks,allcolors=cvprblue]{hyperref}

\title{Dynamic Model Merging Made Slim}

\author{%
  Guodong Du \qquad Wanyu Lin\thanks{Corresponding author.}\\
  Department of Computing, The Hong Kong Polytechnic University\\
  \texttt{guodong.du@connect.polyu.hk} \quad \texttt{wan-yu.lin@polyu.edu.hk}
}

\begin{document}
\maketitle
\suppressfloats[t] 

\input{sec/0_abstract}
\input{sec/1_intro}

\input{sec/2_related_work}
\input{sec/3_backg_motivation}
\input{sec/4_method}
\input{sec/5_experiments}
\input{sec/6_analysis}
\FloatBarrier
\Needspace{24\baselineskip}
\input{sec/7_conclusion}
\par
\begingroup
\parfillskip=0pt plus 0.5\linewidth
\input{camera_ready_disclosures}
\endgroup

\clearpage
\bibliography{main}

\newpage
\appendix
\input{sec/8_appendix}

\newpage
\input{sec/9_rebuttal_integration}

\newpage
\input{checklist}

\end{document}

%% file: preamble.tex
\newcommand{\ourapproach}{\texttt{DiDi-Merging}\xspace}
\newcommand{\foo}[2]{$\text{#1}\ _{\textcolor{darkblue}{(\text{#2})}}$}
\newcommand{\pub}[1]{{\color{gray}{\tiny{[{#1}]\!}}}}

\definecolor{mygray}{gray}{.9}
\definecolor{mygray2}{HTML}{F5F5F5}
\definecolor{mypink}{HTML}{FDF5F1}
\definecolor{darkred}{RGB}{156, 39, 33}
\definecolor{dark1}{gray}{0.85}
\definecolor{tabhighlight}{HTML}{e5e5e5}
\definecolor{darkblue}{RGB}{31, 90, 153}
\definecolor{mylightblue}{rgb}{0.8, 0.9, 1.0}
\definecolor{mylightgreen}{rgb}{0.8, 1.0, 0.9}
\definecolor{mylightred}{rgb}{1.0, 0.9, 0.8}
\definecolor{darkgreen}{rgb}{0,0.5,0}
\definecolor{azureblue}{rgb}{0,0.5,1}
\definecolor{color1}{HTML}{006EB8}
\definecolor{color2}{HTML}{009B55}
\definecolor{color3}{HTML}{00A99A}
\definecolor{color4}{HTML}{3C8031}
\definecolor{forestGreen}{RGB}{34, 139, 34}
\definecolor{firebrick}{RGB}{178, 34, 34}
\definecolor{cvprblue}{rgb}{0.21,0.49,0.74}

\newcommand{\myxmark}{\color{firebrick}{\ensuremath{\times}}}
\newcommand{\mycheckmark}{\color{forestGreen}{\checkmark}}

\usepackage{titletoc}
\usepackage{tabularx}
\usepackage{dsfont}
\usepackage{colortbl}
\usepackage{booktabs}
\usepackage{multirow}
\usepackage{amsmath, amsfonts, amssymb}
\usepackage{makecell}
\usepackage{arydshln}
\usepackage{nicematrix}
\usepackage{adjustbox}
\usepackage{flafter}
\usepackage{placeins}
\usepackage{subcaption}
\usepackage{xspace}
\usepackage{graphicx}
\usepackage{pifont}     


%% file: sec/0_abstract.tex
\begin{abstract}
Model merging consolidates independently fine-tuned checkpoints without
joint training or access to task data. Dynamic methods retain task-specific
capacity, but must divide a limited storage budget between shared and
expert parameters. We propose DiDi-Merging, which jointly allocates
low-rank capacity to these components through differentiable rank
selection, followed by data-free parameter-space refinement. On the
eight-task ViT-B/32 benchmark, the small configuration reaches $89.3\%$
average accuracy ($98.7\%$ normalized retention) at $1.24\times$ the
storage of one backbone, including that backbone; the medium
configuration reaches $89.8\%$ at $1.4\times$. Removing refinement
reduces the medium configuration by $4.3$ percentage points, while
static-merge plus uniform-recovery controls are $4.7$--$8.9$ points below
the small configuration. Additional evaluations cover language and
multimodal models at their separately reported budgets. The main results
assume known task identity; practical routing and accuracy under larger
task counts at fixed capacity remain distinct evaluation questions. Code is
available at \url{https://github.com/WanyuGroup/NeurIPS2026-DiDi-Merging}.
\end{abstract}

%% file: sec/1_intro.tex
\section{Introduction}
\label{sec:introduction}

Fine-tuning many task-specific models from a shared
backbone~\cite{radford2021learning,touvron2023llama} is now standard,
but deploying them at scale is expensive, and joint multi-task
training across the original datasets is often infeasible due to
compute or privacy constraints~\cite{weitask}. Model merging
combines fine-tuned models into a single unified model at the
parameter level without access to the original training
data~\cite{jin2022dataless,du2024knowledge}.

\begin{figure}[!t]
  \centering
  \includegraphics[width=0.54\linewidth]{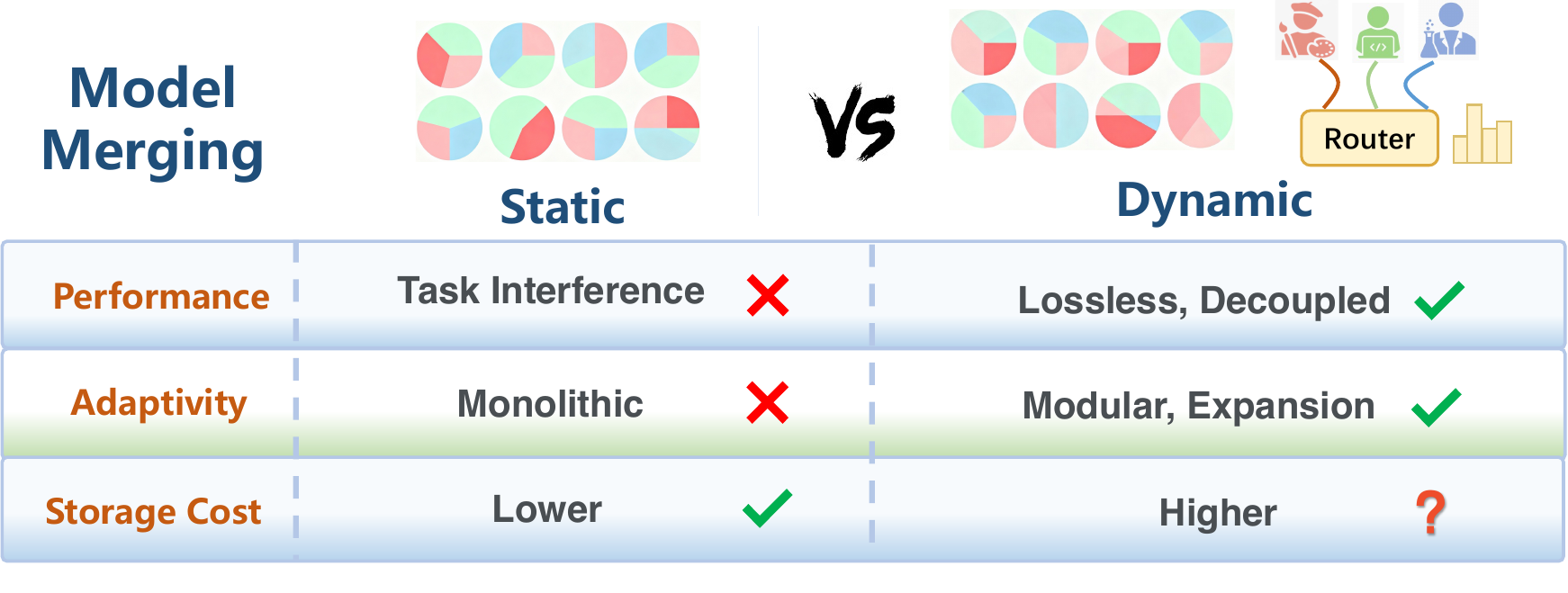}
  \caption{Static and dynamic model merging.}
  \label{fig:static}
\end{figure}
Most existing work focuses on static merging~\cite{ta,ties}, which
fuses fine-tuned models into a single monolithic checkpoint and
suffers from task interference. Recent dynamic merging
methods~\cite{mmer} route inputs to task-relevant expert components,
avoiding interference and supporting modular expansion
(Fig.~\ref{fig:static}). The remaining bottleneck is storage:
dynamic methods pay extra parameter overhead per task, motivating
slim dynamic merging schemes.

\begin{figure}[!t]
  \centering
  \begin{minipage}[t]{0.24\linewidth}
    \centering
    \includegraphics[width=\linewidth]{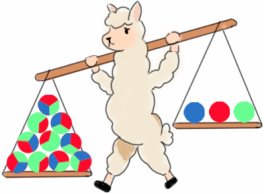}
    \subcaption{Heavy Shared Knowledge}
    \label{fig:effect1}
  \end{minipage}\hfill
  \begin{minipage}[t]{0.24\linewidth}
    \centering
    \includegraphics[width=\linewidth]{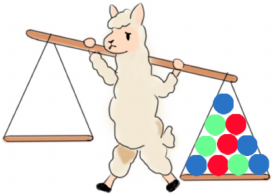}
    \subcaption{Heavy Experts Knowledge}
    \label{fig:effect2}
  \end{minipage}\hfill
  \begin{minipage}[t]{0.24\linewidth}
    \centering
    \includegraphics[width=\linewidth]{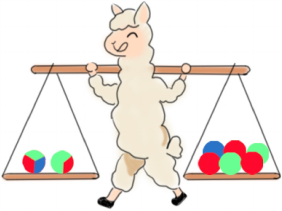}
    \subcaption{Balanced Shared--Expert}
    \label{fig:effect3}
  \end{minipage}\hfill
  \begin{minipage}[t]{0.24\linewidth}
    \centering
    \includegraphics[width=\linewidth]{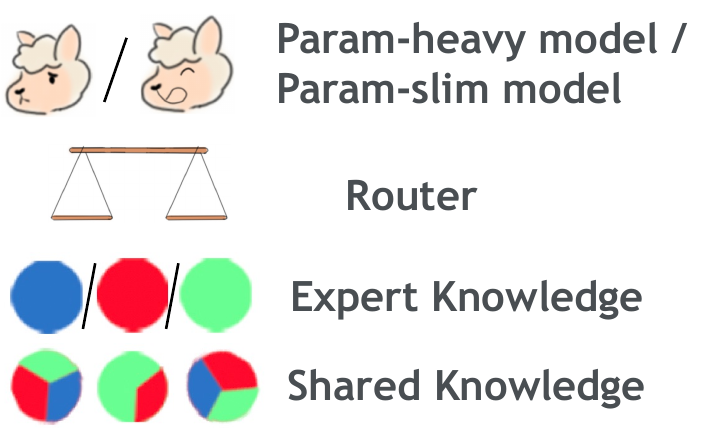}
    \label{fig:effect4}
  \end{minipage}
  \caption{Parameter allocation in dynamic model merging. The balance
  beam denotes the router, and baskets hold shared (multi-colour
  circles) and expert (single-colour circles) knowledge; legend at
  right.
  (a) \emph{Heavy shared}: full-size shared model, tiny per-task
  experts.
  (b) \emph{Heavy expert}: no shared component; per-task expert
  storage grows with task count.
  (c) \emph{Balanced shared--expert}: slim shared module $+$ small
  per-task experts, the regime \ourapproach\ targets via
  differentiable rank optimisation, with capacity adapting to task
  similarity.}
  \label{fig:hyperparams10}
\end{figure}

Existing dynamic merging methods exhibit two failure modes that lie at
opposite ends of the same axis: \emph{how parameters are split between
shared and expert components}. The first
(\emph{heavy shared}, Fig.~\ref{fig:effect1}) keeps a full-size shared
model and stores only tiny per-task expert vectors:
TALL-Mask~\cite{talls}, EMR-Merging~\cite{emr},
Twin-Merging~\cite{twin}, FREE-Merging~\cite{free}, and MMER~\cite{mmer}
all fall into this regime, with parameters concentrated in a shared
component no smaller than the original model. The second
(\emph{heavy expert}, Fig.~\ref{fig:effect2}) does the opposite:
TSV-C~\cite{tsv} and SMILE~\cite{smile} drop the shared component and
push all task knowledge into per-task expert vectors, so the total
expert storage grows with the number of tasks even when each vector is
low-rank. Neither extreme balances the two
(Fig.~\ref{fig:effect3}).
To address these limitations, we propose \textbf{DiDi}-Merging
(\textbf{D}ynam\textbf{i}c Merging with \textbf{Di}fferentiable Rank), which
balances parameters between shared and task-specific experts, yielding a slim
and efficient dynamic merging framework.

\textbf{DiDi}-Merging proceeds in three steps. (1) Task vectors are
statically merged into a shared knowledge vector, with per-task
residuals as expert vectors (exact but parameter-heavy). (2) Low-rank
decomposition recasts parameter budgeting as a differentiable rank
allocation problem, learned end-to-end against a task-vector
reconstruction objective. (3) The truncated low-rank parameters are
refined in a data-free pass to recover task-specific performance.
Unlike prior differentiable-SVD methods for compressing a
\emph{single} model~\cite{dobisvd}, our \emph{cross-task} rank
allocation jointly optimizes shared and expert components against
task-vector reconstruction targets. The full pipeline runs in two
stages (rank optimisation followed by data-free LoRA refinement),
illustrated in Fig.~\ref{fig:method}.

On the eight-task ViT-B/32 benchmark, \ourapproach-S reaches $89.3\%$
average accuracy ($98.7\%$ normalized retention) at $1.24\times$ the
storage of one backbone; \ourapproach-M reaches $89.8\%$ at $1.4\times$.
The language and multimodal evaluations use their separately reported
budgets and metrics, rather than inheriting the vision operating point.
The method combines a shared--expert decomposition, coupled rank
allocation, and parameter-space refinement. Its contribution is this
budgeted consolidation formulation and the resulting evaluated trade-off,
not the invention of SVD, LoRA, or differentiable rank selection.

This paper makes three \textbf{contributions}:
\begin{itemize}
    \item We identify two failure modes in existing dynamic merging
    methods (\emph{heavy shared} and \emph{heavy expert}), both
    stemming from one-sided shared/expert allocation.
    \item We propose \textbf{DiDi}-Merging, a
    \textbf{d}ynam\textbf{i}c merging framework with
    \textbf{di}fferentiable rank allocation: ranks are jointly
    optimized for shared and per-task expert components against
    task-vector reconstruction targets, without access to original
    training data.
    \item We evaluate accuracy and storage across vision, language, and
    multimodal settings, with controlled comparisons that separate the
    value of adaptive rank allocation from that of uniform-rank recovery.
\end{itemize}

\input{tabs/intro}

%% file: tabs/intro.tex
\begin{figure}[!t]
\centering
\begin{minipage}[t]{0.40\linewidth}
\vspace{0pt}
    \centering
    \includegraphics[width=\linewidth]{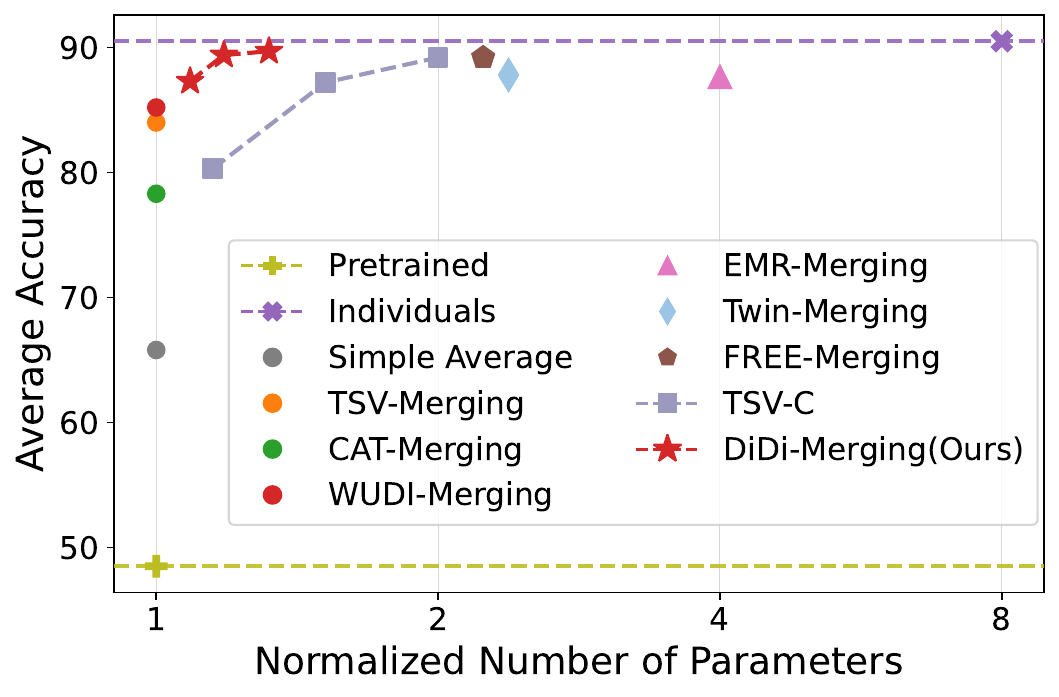}
    \captionof{figure}{\textbf{DiDi}-Merging achieves a superior
    Pareto trade-off between parameter efficiency and accuracy across
    all dynamic merging baselines.}
    \label{fig:result}
\end{minipage}\hfill
\begin{minipage}[t]{0.58\linewidth}
\vspace{0pt}
    \centering
    \captionof{table}{Model merging approaches and requirements.}
    \label{tab:comparison}
    \belowrulesep=1.5pt
    \aboverulesep=1.5pt
    \small
    \setlength{\tabcolsep}{6pt}
    \resizebox{\linewidth}{!}{
    \begin{tabular}{ccccc}
        \toprule
        \rowcolor{mygray} &  & \textbf{Data} & \textbf{Slim} & \textbf{Slim} \\
        \rowcolor{mygray} \multirow{-2}{*}{\textbf{Method}}
        & \multirow{-2}{*}{\textbf{Parameters}}
        & \textbf{Free}
        & \textbf{Shared}
        & \textbf{Experts} \\
        \midrule
        \rowcolor{mygray2} Multiple Models \cite{modelsoups} & 8\,X & - & - & - \\
        \rowcolor{mygray2} Traditional MTL \cite{chen2024multi} & 1\,X & \myxmark & - & - \\
        \rowcolor{mygray2} Static Merging \cite{yang_survey} & 1\,X & \mycheckmark & - & - \\
        \midrule
        \multicolumn{5}{c}{\emph{Dynamic Merging Methods}} \\
        WEMoE \cite{wemoe} & 6.27\,X & \myxmark & \mycheckmark & \myxmark \\
        TALL-Mask \cite{talls} & 2.25\,X & \mycheckmark & \myxmark & \mycheckmark \\
        EMR-Merging \cite{emr} & 2.25\,X & \mycheckmark & \myxmark & \mycheckmark \\
        Twin-Merging \cite{twin} & 2.25\,X & \mycheckmark & \myxmark & \mycheckmark \\
        Free-Merging \cite{free} & 2.16\,X & \mycheckmark & \myxmark & \mycheckmark \\
        TSV-C \cite{tsv} & 2.08\,X & \mycheckmark & \mycheckmark & \myxmark \\
        SMILE \cite{smile} & 3.07\,X & \myxmark & \mycheckmark & \myxmark \\
        \rowcolor{mypink} \textbf{DiDi-Merging (Ours)} & \textbf{1.24}\,X & \mycheckmark & \mycheckmark & \mycheckmark \\
        \bottomrule
    \end{tabular}
    }
\end{minipage}
\end{figure}

%% file: sec/2_related_work.tex
\section{Related Work}\label{sec:related_work}

\paragraph{Dynamic Model Merging.}
Recent surveys organize model fusion across parameter, representation,
and behavior levels~\cite{cai2026fusion,zhou2025democratizing,
zhou2026model}; behavior-level alternatives include preference-based
fusion and graph-on-logits distillation~\cite{gu2025infifpo,
wang2025infigfusion}. Static merging~\cite{ta,ties,pcb,nps,fang2025see} fuses
fine-tuned models into a single checkpoint and suffers from task
interference. Dynamic methods route inputs to task-specific experts
to avoid interference~\cite{wemoe,yang_survey,du2025graftllm,li2025multi,
li2026skillweave}, but their parameter allocation
is fixed: \emph{heavy shared} variants
(TALL-Mask~\cite{talls}, EMR-Merging~\cite{emr},
Twin-Merging~\cite{twin}, FREE-Merging~\cite{free},
MMER~\cite{mmer}) keep a full-size shared model with masks or
rescalers to identify task-relevant subspaces, while \emph{heavy
expert} variants (TSV-C~\cite{tsv}, SMILE~\cite{smile}) drop the
shared component and compress per-task vectors via SVD or subspace
modelling. Recent methods also construct reusable component-level expert
libraries for input-aware recombination~\cite{qiu2026merge} or evolve
experts through manifold-aware, training-free routing for continual
merging~\cite{qiu2026madeit}. Complementary work makes expert training
merge-aware~\cite{gu2026smat}, calibrates or quantizes post-merge
models~\cite{gu2026featcal,wang2026epmq}, and studies merging scaling
laws or expert-read budgets~\cite{wang2026modelmergingscaling,
wang2026mergepipe}. \ourapproach\ instead \emph{optimizes} this allocation via
a differentiable rank objective; full discussion in
App.~\ref{app:a}.

\paragraph{Task Vector Compression and Differentiable Rank Selection.}
Most merging methods build on task-vector
arithmetic~\cite{ta,zhang2023composing,ties,dare,pcb,infifusion} or
further compress task vectors via
masking~\cite{grafting,mask,tailor},
SVD~\cite{wang2024svd,fang2025see,impart,knots,twin,delta-moe}, or
quantization~\cite{frantar2022gptq,lin2024awq,bitdelta,delta-come,li2026d},
yet they allocate storage uniformly across tasks and overlook the
shared/expert distinction.
Differentiable rank selection has been studied outside
merging: AdaLoRA~\cite{adalora} prunes singular values during
fine-tuning, DyLoRA~\cite{dylora} trains a single LoRA across a
range of ranks for inference-time selection, and
Dobi-SVD~\cite{dobisvd} introduces differentiable truncation for
single-model LLM compression. \ourapproach\ differs in three ways:
\emph{(i) Setting:} we operate on multiple pre-existing task
vectors for post-hoc merging rather than fine-tuning or single-model
compression;
\emph{(ii) Data:} we are fully data-free, supervised by the stored
task vectors themselves rather than a calibration set;
\emph{(iii) Allocation scope:} ranks are jointly optimized across
task vectors \emph{and} the shared/expert decomposition, rather than
within a single matrix or model. Related multimodal studies also organize
shared and specialized information: \citet{meng2026dbr} propose DBR, which
rebalances shared and private branches to reduce shared redundancy while
preserving modality-specific evidence.

%% file: sec/3_backg_motivation.tex
\section{Background and Motivation}\label{sec:motivation}

\subsection{Preliminaries}\label{sec:preliminary}
Consider a pre-trained model $f$ with parameters $\boldsymbol{\theta}_0$
and $T$ downstream tasks fine-tuned to $\boldsymbol{\theta}_t$, with
\emph{task vector} $\boldsymbol{\tau}_t = \boldsymbol{\theta}_t -
\boldsymbol{\theta}_0$. \textbf{Static merging} fuses these into a
single input-agnostic model $\boldsymbol{\theta}^* =
\boldsymbol{\theta}_0 + \sum_{t=1}^T \lambda_t \boldsymbol{\tau}_t$,
while \textbf{dynamic merging} activates shared and task-specific
components conditional on the input:
\begin{equation}\label{eq:dynmerge}
    \boldsymbol{\theta}^*(\boldsymbol{x})
    = \boldsymbol{\theta}_0 + \boldsymbol{M}_s
    + \sum_{t=1}^{T} w_t(\boldsymbol{x}) \boldsymbol{M}_t,
\end{equation}
where $\boldsymbol{M}_s$ and $\boldsymbol{M}_t$ encode shared and
expert knowledge, and $w_t(\boldsymbol{x})$ are lightweight,
input-dependent routing weights from a small router. Following prior
dynamic merging work~\cite{twin,mmer}, we use an MoE-style
\emph{ideal router} that takes task identity at inference; routing
design is orthogonal to our contribution, which targets parameter
organisation across shared and expert components. All reported total-storage ratios include one pretrained backbone.
The ideal task-ID selector has no learned router parameters; a learned
router would incur an additional cost that must be counted separately.
The main results assume known task identity and do not evaluate routing
for unknown tasks.

\begin{figure}[!t]
\centering
\begin{minipage}[t]{0.66\linewidth}
\vspace{0pt}
  \centering
  \includegraphics[width=\linewidth]{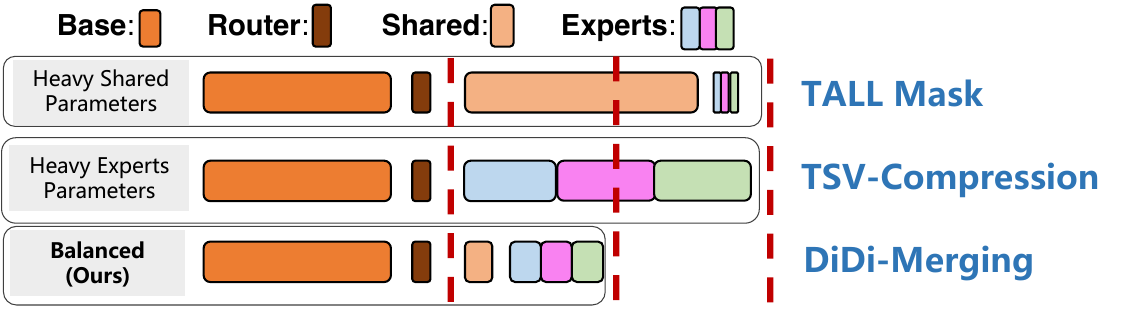}
  \captionof{figure}{Parameter accounting under three dynamic
  merging regimes. \emph{Heavy shared} inflates the shared
  component; \emph{heavy expert} inflates per-task experts;
  \ourapproach\ keeps both small. Decomposition
  $|\boldsymbol{\theta}_0| + |\mathrm{router}| + r_s(m{+}n) + T\,r_t(m{+}n)$
  and the $1.24\times/1.4\times/2.0\times$ budgets in
  App.~\ref{app:storage}.}
  \label{fig:novelty}
\end{minipage}\hfill
\begin{minipage}[t]{0.32\linewidth}
\vspace{0pt}
  \centering
  \includegraphics[width=\linewidth]{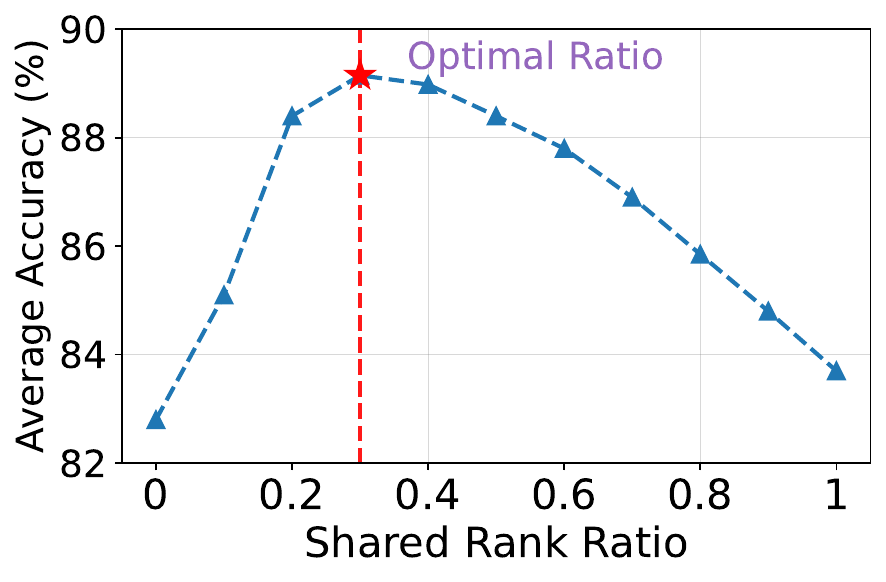}
  \captionof{figure}{Effect of rank allocation on dynamic merging
  accuracy at a fixed parameter budget on the 8-task vision
  benchmark.}
  \label{fig:rethink}
\end{minipage}
\end{figure}
\subsection{Rethinking Dynamic Model Merging}\label{sec:rethink}
\noindent\textbf{Q1: How should shared and expert knowledge be
represented?}
As Fig.~\ref{fig:novelty} shows, prior dynamic methods take one of
two extremes: \emph{heavy shared} (full-size $\boldsymbol{M}_s$,
e.g., \cite{emr,twin,free}) or \emph{heavy expert} (no
$\boldsymbol{M}_s$, e.g., \cite{tsv}). We instead represent
\emph{both} as \textcolor{darkred}{low-rank modules} with
\textcolor{darkred}{flexible rank allocation}: similar tasks rely on
shared ranks, diverse tasks on expert ranks
(Fig.~\ref{fig:heatmap}).

\noindent\textbf{Q2: How to allocate ranks across shared and expert
modules?}
At a fixed parameter budget, the choice of ranks substantially
affects accuracy (Fig.~\ref{fig:rethink}). We introduce a
\textcolor{darkred}{differentiable rank mechanism}: each weight
matrix is SVD-decomposed and its singular values are reweighted by a
\textcolor{darkred}{smooth truncation function} controlled by a
learnable continuous rank $r$. Jointly minimising a task
reconstruction loss and a compression penalty yields a
\textcolor{darkred}{slim allocation} for each module.

\noindent\textbf{Q3: How to recover task fidelity after rank
truncation?}
Direct truncation degrades accuracy, so we apply a
\textcolor{darkred}{data-free optimization} that refines the
low-rank module parameters using the stored task vectors themselves
as \textcolor{darkred}{reconstruction targets}.

%% file: sec/4_method.tex
\section{Method}
\label{sec:method}
\begin{figure}[!t]
\centering
\includegraphics[width=\linewidth]{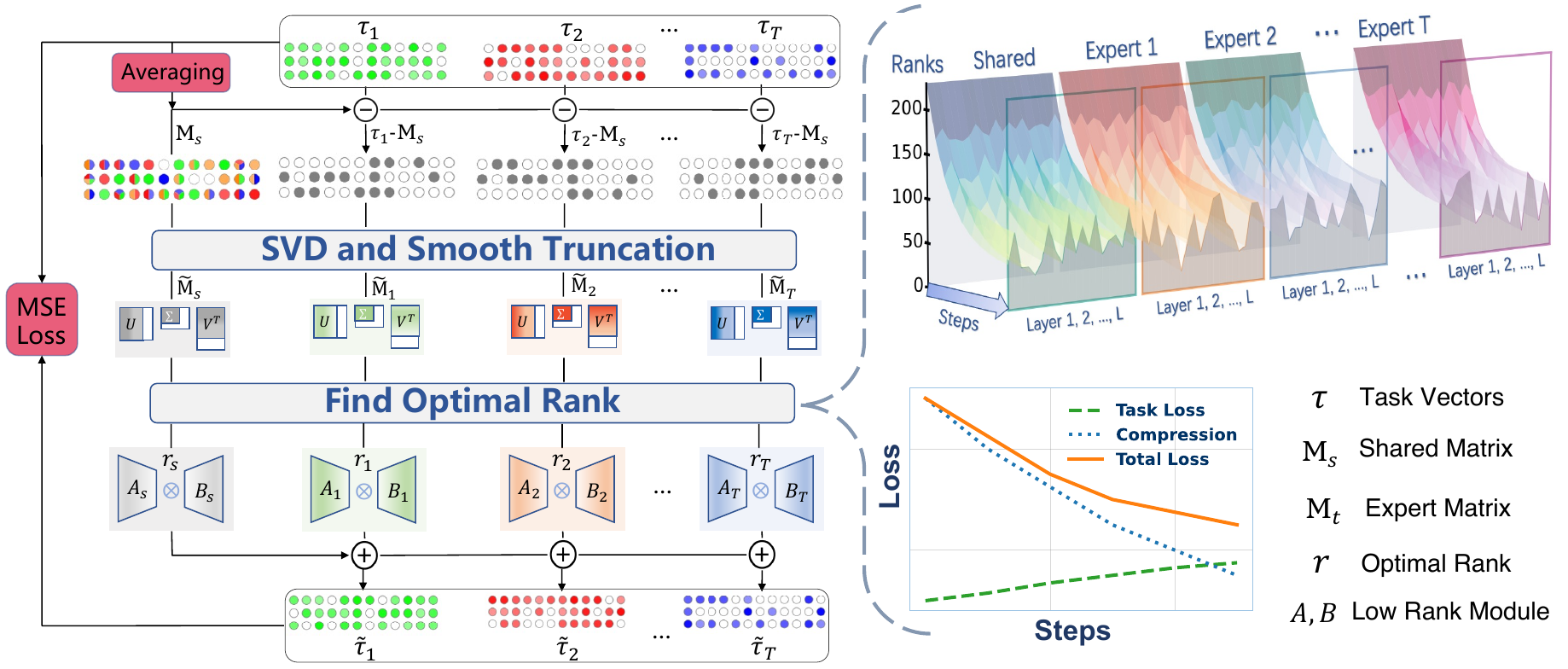}
\caption{The \textbf{DiDi}-Merging pipeline.
\emph{Left:} task vectors $\{\boldsymbol{\tau}_t\}$ are split into a
shared component $\boldsymbol{M}_s$ via averaging and per-task
expert residuals $\boldsymbol{\tau}_t - \boldsymbol{M}_s$; SVD with
smooth truncation produces low-rank $\tilde{\boldsymbol{M}}_s,
\tilde{\boldsymbol{M}}_t$, the Find-Optimal-Rank stage learns
per-module ranks $r_s, r_1,\dots,r_T$, and LoRA factors $(A,B)$ are
refined data-free against an MSE reconstruction loss with respect
to the original task vectors. \emph{Right:} during optimisation,
per-layer ranks evolve separately for shared and expert modules
(top), while the joint objective trades task reconstruction against
compression (bottom).}
\label{fig:method}
\end{figure}

We present \textbf{DiDi}-Merging, a slim dynamic merging framework
(Fig.~\ref{fig:method}).
Our approach consists of three key steps: establishing shared and expert
matrices (\S\ref{subsec:method1}), differentiable optimization of ranks to
determine optimal allocations under parameter constraints
(\S\ref{subsec:method2}), and a data-free refinement that restores
performance lost from rank truncation (\S\ref{subsec:method3}).

\subsection{Low-Rank Decomposition}
\label{subsec:method1}
Building upon the task vector representation introduced in the preliminary
section, we decompose both shared and expert knowledge into low-rank
modules to enable efficient dynamic merging.
The shared matrix $\boldsymbol{M}_s$ is obtained via static merging of
the task vectors $\{\boldsymbol{\tau}_t\}_{t=1}^{T}$, e.g., by
averaging:
\begin{equation}
\boldsymbol{M}_s = \mathrm{avg}(\boldsymbol{\tau}_1, \dots, \boldsymbol{\tau}_T),
\end{equation}
and each expert matrix is defined as the residual relative to the
shared component:
\begin{equation}
\boldsymbol{M}_t = \boldsymbol{\tau}_t - \boldsymbol{M}_s.
\end{equation}

To achieve controllable and efficient parameterization, we apply singular
value decomposition (SVD):
\begin{equation}
\boldsymbol{M}_s = \boldsymbol{U}_s \boldsymbol{\Sigma}_s \boldsymbol{V}_s^{\top}, \quad
\boldsymbol{M}_t = \boldsymbol{U}_t \boldsymbol{\Sigma}_t \boldsymbol{V}_t^{\top}.
\end{equation}
Retaining only the top-$r$ singular components yields the low-rank modules:
\begin{equation}
\tilde{\boldsymbol{M}}_s^{(r)} = \boldsymbol{U}_s^{(r)} \boldsymbol{\Sigma}_s^{(r)} {\boldsymbol{V}_s^{(r)}}^\top, \quad
\tilde{\boldsymbol{M}}_t^{(r)} = \boldsymbol{U}_t^{(r)} \boldsymbol{\Sigma}_t^{(r)} {\boldsymbol{V}_t^{(r)}}^\top.
\end{equation}
Finally, the dynamically merged parameters are
\begin{equation}
\boldsymbol{\theta}^*(\boldsymbol{x})
= \boldsymbol{\theta}_0
+ \tilde{\boldsymbol{M}}_s^{(r)}
+ \sum_{t=1}^{T} w_t(\boldsymbol{x}) \cdot \tilde{\boldsymbol{M}}_t^{(r)}.
\end{equation}

\subsection{Differentiable Optimization of Ranks}\label{subsec:method2}
To efficiently determine the optimal rank of each shared and expert module
in a layer, we leverage a differentiable approach based on the smooth
truncation of the singular values from the SVD of the matrix
$\boldsymbol{M} \in \mathbb{R}^{m \times n}$, $m > n$:
\begin{equation}
\boldsymbol{M} = \boldsymbol{U} \boldsymbol{\Sigma} \boldsymbol{V}^\top, \quad
\boldsymbol{\Sigma} = \mathrm{diag}(\sigma_1, \sigma_2, \dots, \sigma_n),
\end{equation}
where $\boldsymbol{U} \in \mathbb{R}^{m \times n}$ and
$\boldsymbol{V} \in \mathbb{R}^{n \times n}$.
We define a smooth truncation function to retain a differentiable subset of
singular values:
\begin{equation}
    \tilde{\sigma}_i = \sigma_i \left[0.5 \cdot \tanh\big(\beta (r - i)\big) + 0.5 \right],
\end{equation}
where $r$ is a learnable continuous rank parameter, and $\beta$ controls
the smoothness of the truncation.
Comparison with AdaLoRA~\cite{adalora} and
Dobi-SVD~\cite{dobisvd} is in \S\ref{sec:related_work} and
App.~\ref{app:g}.
The truncated low-rank matrix is reconstructed as
\begin{equation}
    \tilde{\boldsymbol{M}} = \boldsymbol{U} \,
        \mathrm{diag}(\tilde{\sigma}_1, \tilde{\sigma}_2, \dots, \tilde{\sigma}_n) \,
        \boldsymbol{V}^\top.
\end{equation}
During optimization, only the rank variable $r$ is updated via gradient
descent, while other low-rank module parameters remain frozen.

\paragraph{Task Reconstruction Loss.}
The reconstructed per-layer task vector is given by
\begin{equation}
    \tilde{\boldsymbol{\tau}}_{t,l}
    = \tilde{\boldsymbol{M}}_{s,l} + \tilde{\boldsymbol{M}}_{t,l},
\end{equation}
where $\tilde{\boldsymbol{M}}_{s,l}$ and $\tilde{\boldsymbol{M}}_{t,l}$
denote the truncated shared and expert matrices at layer $l$.
The task reconstruction loss is the mean squared error over all
tasks and layers:
\begin{equation}
    L_{\text{task}} = \frac{1}{T} \sum_{t=1}^{T} \sum_{l=1}^{L}
    \big\| \tilde{\boldsymbol{\tau}}_{t,l} - \boldsymbol{\tau}_{t,l} \big\|_F^2,
\end{equation}
where $T$ is the number of tasks and $L$ the number of layers.

\paragraph{Coupled parameter budget.}
For a layer of shape $m_l\times n_l$, storing both shared factors and
all expert factors costs
\begin{equation}
 B_{\rm factors}=\sum_l(m_l+n_l)\left(r_{s,l}+\sum_{t=1}^{T}r_{t,l}\right).
\label{eq:coupled_budget}
\end{equation}
The rank penalty controls this total allocation, rather than granting
an independent full budget to each expert. $R_{\rm now}$ denotes the
factor count normalized by the corresponding dense matrix count;
$R_{\rm tar}$ sets its target. The reported deployment ratio additionally
includes the pretrained backbone and any router, as detailed in
Appendix~\ref{app:storage}. Shared and expert factors compete for the
same storage budget, and integer ranks must be counted after truncation.

\paragraph{Total Loss.}
The objective balances task reconstruction performance and compression
ratio:
\begin{equation}
    L = L_{\text{task}} + \gamma \cdot \big| R_{\text{now}} - R_{\text{tar}} \big|,
\end{equation}
where $L_{\text{task}}$ is the task reconstruction loss, $R_{\text{tar}}$
the target compression ratio, and $\gamma$ a weighting factor.

\subsection{Low-rank Module Update}
\label{subsec:method3}
After obtaining the optimal rank allocation $\{r_{t,l}\}$ for each task
$t$ and layer $l$, we instantiate low-rank LoRA-style modules to recover
performance.
For each layer $l$ and task $t$, we introduce factor matrices
\[
\boldsymbol{A}_{t,l}\in\mathbb{R}^{m_l\times r_{t,l}},\qquad
\boldsymbol{B}_{t,l}\in\mathbb{R}^{r_{t,l}\times n_l},
\]
and express the low-rank module parameter update as
\begin{equation}
\boldsymbol{M}_{t,l} = \boldsymbol{A}_{t,l}\boldsymbol{B}_{t,l}.
\end{equation}

\paragraph{Initialization.}
To provide a good starting point and accelerate convergence, we
initialize each pair $(\boldsymbol{A}_{*,l}, \boldsymbol{B}_{*,l})$ for
$* \in \{s, 1, \dots, T\}$ (one shared module and $T$ per-task experts)
using the truncated SVD of the precomputed matrix
$\boldsymbol{M}_{*,l}^{(0)}$:
\[
\boldsymbol{M}_{*,l}^{(0)} \approx \boldsymbol{U}_{*,l}^{(r)} \boldsymbol{\Sigma}_{*,l}^{(r)} {\boldsymbol{V}_{*,l}^{(r)}}^\top,
\]
We initialize the factors as
\[
\boldsymbol{A}_{*,l} \gets \boldsymbol{U}_{*,l}^{(r)} (\boldsymbol{\Sigma}_{*,l}^{(r)})^{1/2}, \qquad
\boldsymbol{B}_{*,l} \gets (\boldsymbol{\Sigma}_{*,l}^{(r)})^{1/2} {\boldsymbol{V}_{*,l}^{(r)}}^\top,
\]
so that $\boldsymbol{A}_{*,l} \boldsymbol{B}_{*,l} \approx \boldsymbol{M}_{*,l}^{(0)}$.

\paragraph{Data-free Task Reconstruction.}
We supervise using the existing task vectors $\{\boldsymbol{\tau}_t\}$.
For task $t$, the reconstructed vector is computed by summing the
contributions of the low-rank shared and task-specific LoRA modules across
layers:
\begin{equation}
\tilde{\boldsymbol{\tau}}_{t,l} = \boldsymbol{A}_{s,l}\boldsymbol{B}_{s,l}
+ \boldsymbol{A}_{t,l}\boldsymbol{B}_{t,l},
\end{equation}
where $\boldsymbol{A}_{s,l}\boldsymbol{B}_{s,l}$ and
$\boldsymbol{A}_{t,l}\boldsymbol{B}_{t,l}$ denote shared and task-specific
LoRA modules, respectively.
The task reconstruction loss is then defined as the mean squared error
over all tasks and layers:
\begin{equation}
L_{\text{task}} = \frac{1}{T} \sum_{t=1}^{T} \sum_{l=1}^{L}
\big\| \tilde{\boldsymbol{\tau}}_{t,l} - \boldsymbol{\tau}_{t,l} \big\|_F^2.
\end{equation}

\paragraph{Optimization Protocol.}
In this stage, the shared and task-specific factor matrices are
refined with the ranks fixed, while the pretrained backbone remains frozen.
We perform a lightweight, data-free optimization using Adam to minimize
$L_{\text{task}}$ (more optimization details in App.~\ref{app:g}).

%% file: sec/5_experiments.tex
\section{Experiments}
\label{sec:exps}
We evaluate vision (\S\ref{exp:vision}), LLM (\S\ref{exp:llm}), and
MLLM (\S\ref{exp:mllm}) merging, with zero-shot and separate fine-tuned
models as references. Total-storage ratios include one backbone and all
shared and expert parameters; the task-ID selector has no learned router.
We report accuracy or the specified benchmark score. Baseline, dataset,
and training details appear in Appendices~\ref{app:d}--\ref{app:f2}.

The evaluated budgets are $1.24\times$ (S), $1.4\times$ (M), and
$2\times$ (L), set by scaling the optimized per-layer ranks. Vision uses
S/M, while LLM and MLLM use L. These are the evaluated operating points;
relative storage alone does not isolate the effects of rank allocation
and refinement. PEFT comparisons and additional modality results are in
Appendices~\ref{app:peft_comparison} and~\ref{app:c}.

\subsection{Merging Vision Models}
\label{exp:vision}
\input{tabs/vision_result1}
\input{tabs/vision_result2}
\input{tabs/llm}
\textbf{Setup.}
Following Ilharco et al.~\cite{ta,pcb}, we evaluate on
CLIP-ViT-B/32 and ViT-L/14~\cite{radford2021learning} across eight
classification benchmarks (SUN397, Cars, RESISC45, EuroSAT, SVHN,
GTSRB, MNIST, DTD; details in App.~\ref{app:e}), and compare
against five static and seven dynamic baselines listed in
Tab.~\ref{tab:vision1}.

\noindent\textbf{Vision Results.}
Tab.~\ref{tab:vision1} reports accuracy and parameter overhead on
the 8-task suite. Static methods add no storage but trail dynamic
baselines by $4$--$6\%$ in accuracy, with the strongest static
method (WUDI-Merging at $85.2\%$) still $4.3\%$ below the best
dynamic baseline. At $1.24\times$, \ourapproach-S matches the
strongest dynamic baseline ($89.3\%$ vs.\ TSV-C at $89.5\%$) using
under $60\%$ of their storage; at $1.4\times$, \ourapproach-M
attains the highest score overall ($89.8\%$), surpassing all
baselines. Scaling to 14 / 20 tasks and ViT-L/14
(Tab.~\ref{tab:vision2}), \ourapproach-M leads at 14 / 20 tasks
($99.5\% / 99.4\%$ retention) and ties TSV-C within $0.06\%$ on
ViT-L/14 at less than half the overhead. These evaluations cover different backbone sizes and task counts,
with the corresponding settings and operating points reported in the tables.

\subsection{Merging Large Language Models}
\label{exp:llm}
\textbf{Setup.} We merge four fine-tuned LLMs derived from
Llama-3.1-8B-Instruct~\cite{dubey2024llama} and Qwen-2.5-7B-Instruct~\cite{bai2023qwen},
each obtained by sequential SFT and DPO on a different task
domain (general, mathematics, code, and instruction
following), following FuseLLM~\cite{fusellm} and
FuseChat~\cite{fusechat3}. Training configurations are in
App.~\ref{app:f1}. We compare against state-of-the-art dynamic
baselines Twin-Merging and FREE-Merging.

\noindent\textbf{Datasets.} Evaluation spans ten benchmarks across
four categories: general (MMLU-Pro, MMLU-Redux, GPQA-Diamond),
mathematics (GSM8K, MATH, AMC23), code (HumanEval, MBPP), and
instruction following (AlpacaEval-2, MT-Bench). Details in
App.~\ref{app:e}.

\noindent\textbf{LLM Results.} LLM merging is harder than vision
(more diverse capabilities, larger task heterogeneity); we report
\ourapproach-L ($2\times$) for parity with prior dynamic baselines
that all use $\geq 2\times$ storage (Tab.~\ref{tab:llm_result}).
\ourapproach\ consistently outperforms prior methods, particularly
on challenging benchmarks like AlpacaEval-2, where the gains over FREE-Merging are $5.4$ and $5.1$ points for
Llama-3.1-8B-Instruct and Qwen-2.5-7B-Instruct, respectively.
The table-wide averages increase by $1.8$ and $1.4$ points.

\subsection{Merging Multimodal Large Language Models}
\label{exp:mllm}
\input{tabs/mllm}
\textbf{Setup.}
We merge four MLLMs (Vision, Audio, Video, Point Cloud), each
fine-tuned from Vicuna-7B-v1.5~\cite{51} under identical
environments following~\cite{18,20}; configurations are in
App.~\ref{app:f2}. We compare against representative data-free
dynamic baselines NaiveMC~\cite{18} and MMER~\cite{mmer},
reporting accuracy/scores and retention relative to each fine-tuned
counterpart.

\noindent\textbf{Datasets.}
We use fourteen dual-modal tasks, each pairing text with another
modality: vision (VQAv2~\cite{56}, GQA~\cite{57},
TextVQA~\cite{59}, VizWiz~\cite{60}, ScienceQA~\cite{61},
POPE~\cite{62}, OK-VQA~\cite{63}), audio (TUT~\cite{64},
VocalSound~\cite{65}, Clotho~\cite{66}), video
(MSRVTT~\cite{67}, MSVD~\cite{68}), and point cloud
(ModelNet40~\cite{55}, Objaverse~\cite{58}).

\noindent\textbf{MLLM Results.}
Tab.~\ref{tabs:mllm_result} shows \ourapproach\ consistently
outperforms prior dynamic methods at a slim parameter footprint;
NaiveMC's efficient merging suffers substantial performance
degradation. Notably, all dynamic methods (including ours) exceed
the fine-tuned baseline on several audio and point-cloud tasks: this is consistent with useful transfer across the evaluated models.
The benchmark scores do not by themselves identify a specific transfer
mechanism; additional analysis is in App.~\ref{app:b3}.

%% file: tabs/vision_result1.tex
\begin{table}[!t]
\centering
\belowrulesep=0pt
\aboverulesep=0pt
\caption{Test set performance when merging ViT-B/32 models on 8 vision
tasks, with normalized averages shown in parentheses.}
\label{tab:vision1}
\resizebox{\linewidth}{!}{
\begin{tabular}{c|c|c|cccccccc}
\toprule
\rowcolor{mygray}\textbf{Task($\rightarrow$)} &  &   & \multicolumn{8}{c}{\textbf{Test Set Performance}} \\
\cline{4-11}
\rowcolor{mygray}\textbf{Method($\downarrow$)} & \multirow{-2}{*}{\textbf{$\#$Params\textcolor{darkred}{($\downarrow$)}}} & \multirow{-2}{*}{\textbf{Average$_{\textcolor{darkblue}{(\%)}}$\textcolor{darkred}{($\uparrow$)}}}  & SUN397  &  Cars  &  RESISC45  &  EuroSAT  &  SVHN  &  GTSRB  &  MNIST & DTD \\
\midrule
\rowcolor{mygray2}
Individuals & x8  & 90.5$_{\textcolor{darkblue}{(100)}}$  &  75.3  &  77.7  &  96.1  &  99.7  &  97.5  &  98.7  &  99.7  &  79.4 \\
\rowcolor{mygray2}
Multitask\cite{sanhmultitask} & x1  & 88.9$_{\textcolor{darkblue}{(99.2)}}$  &  74.4  &  77.9  &  98.2  &  98.9  &  99.5  &  93.9  &  72.9  &  \cellcolor[HTML]{F5F5F5}95.8 \\
\midrule
\multicolumn{11}{c}{\emph{Static Merging Methods}} \\
Weight Averaging\pub{ICLR23} \cite{ta} & x1 & 65.8$_{\textcolor{darkblue}{(72.7)}}$  &  65.3  &  63.4  &  71.4  &  71.7  &  64.2  &  52.8  &  87.5  &  50.1 \\
TSV-Merging\pub{CVPR25} \cite{tsv}  & x1 & 84.0$_{\textcolor{darkblue}{(92.8)}}$	& 69.1	& 70.7	& 85.5	& 94.3	& 92.0	& 91.9	& 99.3	&  69.2 \\
CAT-Merging\pub{ICML25} \cite{cat}  & x1 & 78.3$_{\textcolor{darkblue}{(86.5)}}$  &  68.1  &  65.4  &  80.5  &  89.5  &  85.5  &  78.5  &  98.6  &  60.7 \\
ISO-CTS\pub{ICML25} \cite{iso}  & x1 & 81.4$_{\textcolor{darkblue}{(89.9)}}$	& 74.4	& 74.4	& 87.2	& 90.4	& 76.8	& 83.3	& 97.4	&  67.0 \\
WUDI-Merging\pub{ICML25} \cite{wudi}  & x1 & 85.2$_{\textcolor{darkblue}{(94.1)}}$	& 71.1	& 71.0	& 85.7	& 95.6	& 94.2	& 94.7	& 99.5	&  69.7 \\
\midrule
\multicolumn{11}{c}{\emph{Dynamic Merging Methods}} \\
WEMoE\pub{ICML24} \cite{wemoe}  & x6.27 & 89.2$_{\textcolor{darkblue}{(98.6)}}$ &  73.7  &  76.8   &  93.4  &  98.2  &  \underline{96.8}  &  98.2  &  \textbf{99.6}  &  \underline{76.6} \\
SMILE\pub{TPAMI25} \cite{smile}  & x3.07 & 89.3$_{\textcolor{darkblue}{(98.7)}}$  &  73.6  &  \underline{77.8}  &  92.0  &  98.3  &  \textbf{96.9}  &  98.1  &  \textbf{99.6}  &  \textbf{78.1} \\
TALL-Mask+TIES\pub{ICML24} \cite{talls}  & x2.25 & 88.7$_{\textcolor{darkblue}{(98.0)}}$  &  75.8  &  72.5  &  93.2  &  99.3  &  96.8  &  \textbf{98.5}  &  99.1  &  74.4 \\
Twin-Merging\pub{NeurIPS24} \cite{twin}  & x2.25 & 87.8$_{\textcolor{darkblue}{(97.0)}}$  &  73.6  &  71.7  &  92.1  &  99.3  &  95.3  &  97.2  &  99.1  &  74.0 \\
EMR-Merging\pub{NeurIPS24} \cite{emr}  & x2.25   & 87.7$_{\textcolor{darkblue}{(96.9)}}$  &  74.1  &  72.7  &  91.9  &  99.4  &  95.8  &  96.9  &  99.1  &  72.1 \\
FREE-Merging\pub{ICCV25} \cite{free}  & x2.16 & 89.2$_{\textcolor{darkblue}{(98.6)}}$  &  76.4  &  77.6  &  \underline{93.4}  &  \textbf{99.5}  &  96.3  &  98.2  &  99.5  &  75.4 \\
TSV-C\pub{CVPR25} \cite{tsv}  & x2.08 & \underline{89.5$_{\textcolor{darkblue}{(98.9)}}$}  &  \underline{77.0}  &  77.3  &  93.4  &  99.2  &  96.3  &  \underline{98.3}  &  99.3  &  75.3 \\
\rowcolor{mypink}
\textbf{\ourapproach-S (ours)}  & \textbf{x1.24} & 89.3$_{\textcolor{darkblue}{(98.7)}}$  &  76.9 &  77.2  &  93.2  &  99.3  &  96.5  &  98.1  &  99.2  &  74.1 \\
\rowcolor{mypink}
\textbf{\ourapproach-M (ours)}  & \underline{x1.4} & \textbf{89.8$_{\textcolor{darkblue}{(99.2)}}$}  &  \textbf{77.3} &  \textbf{77.9}  &  \textbf{93.5}  &  \textbf{99.5}  &  96.4  &  98.2  &  \underline{99.5}  &  76.2 \\
\bottomrule
\end{tabular}
}
\end{table}

%% file: tabs/vision_result2.tex
\begin{table}[!t]
\caption{Average accuracy and parameter overhead across dynamic merging
benchmarks for various models and task counts.}
\label{tab:vision2}
\centering
  \resizebox{0.6\linewidth}{!}{
    \begin{tabular}{c|c|cc|c}
      \toprule
      \rowcolor{mygray}\textbf{Task($\rightarrow$)}  &   & \multicolumn{2}{c|}{\textbf{ViT-B/32}} & \multicolumn{1}{c}{\textbf{ViT-L/14}} \\
      \rowcolor{mygray}\textbf{Method($\downarrow$)}& \multirow{-2}{*}{\textbf{$\#$Params}}  & 14 tasks$_{\textcolor{darkblue}{(\%)}}$ & 20 tasks$_{\textcolor{darkblue}{(\%)}}$ & 8 tasks$_{\textcolor{darkblue}{(\%)}}$ \\ \midrule
      \rowcolor{mygray2}\multicolumn{1}{c|}{Individuals} & x14, 20, 8  & 90.88$_{\textcolor{darkblue}{(100)}}$ & 91.37$_{\textcolor{darkblue}{(100)}}$ & 95.81$_{\textcolor{darkblue}{(100)}}$  \\
      \multicolumn{1}{c|}{TALL-Mask+TIES} &  x2.25  & 90.13$_{\textcolor{darkblue}{(99.2)}}$ & 90.91$_{\textcolor{darkblue}{(99.5)}}$ & 93.92$_{\textcolor{darkblue}{(98.0)}}$ \\
      \multicolumn{1}{c|}{EMR-Merging}    &  x2.25    & 89.93$_{\textcolor{darkblue}{(98.9)}}$ & 89.48$_{\textcolor{darkblue}{(97.9)}}$ & 92.73$_{\textcolor{darkblue}{(96.8)}}$  \\
      \multicolumn{1}{c|}{Twin-Merging}   &  x2.25  & 89.54$_{\textcolor{darkblue}{(98.5)}}$   & 89.97$_{\textcolor{darkblue}{(98.5)}}$  & 92.81$_{\textcolor{darkblue}{(96.9)}}$ \\
      \multicolumn{1}{c|}{FREE-Merging}   &  x2.16  & \underline{90.32}$_{\textcolor{darkblue}{(99.4)}}$   & 90.23$_{\textcolor{darkblue}{(98.7)}}$  & 93.78$_{\textcolor{darkblue}{(97.9)}}$ \\
      \multicolumn{1}{c|}{TSV-C} &  x2.08  & 90.29$_{\textcolor{darkblue}{(99.4)}}$ & \underline{90.64}$_{\textcolor{darkblue}{(99.2)}}$  & \textbf{95.68}$_{\textcolor{darkblue}{(99.9)}}$ \\
      \rowcolor{mypink}\multicolumn{1}{c|}{\textbf{\ourapproach-M (ours)}} & \textbf{x1.4} & \textbf{90.42$_{\textcolor{darkblue}{(99.5)}}$} & \textbf{90.84}$_{\textcolor{darkblue}{(99.4)}}$ & \underline{95.62}$_{\textcolor{darkblue}{(99.8)}}$ \\
      \bottomrule
    \end{tabular}
  }
\end{table}

%% file: tabs/llm.tex
\providecommand{\Cwidth}[1]{\relax}
\newcolumntype{C}[1]{>{\centering\let\newline\\\arraybackslash\hspace{0pt}}m{#1}}

\begin{table}[!t]
\caption{Overall results of dynamic merging 4 different LLMs across 10
benchmark tasks. Twin and FREE denote Twin-Merging and Free-Merging.}
\label{tab:llm_result}
\centering
\setlength\tabcolsep{5pt}
\adjustbox{max width=0.92\linewidth}{
\begin{NiceTabular}{@{}cc|C{25pt}C{25pt}C{30pt}C{30pt}C{43pt}|C{25pt}C{25pt}C{30pt}C{30pt}C{43pt}}
\toprule
\multirow{2}{*}{\textbf{\cellcolor{mygray}LLM}} & \multirow{2}{*}{\textbf{\cellcolor{mygray}Benchmark}} & \multicolumn{5}{c}{\textbf{Llama-3.1-8B-Instruct}} & \multicolumn{5}{c}{\textbf{Qwen-2.5-7B-Instruct}}  \\ \cline{3-12}

\cellcolor{mygray} & \cellcolor{mygray} & \rule{0pt}{11pt}\cellcolor{mygray2}Zero-shot & \rule{0pt}{11pt}\cellcolor{mygray2}Fine-tuned & \rule{0pt}{11pt}Twin & \rule{0pt}{11pt}FREE & \rule{0pt}{11pt}\cellcolor{mypink}\textbf{DiDi-Merg.-L} & \rule{0pt}{11pt}\cellcolor{mygray2}Zero-shot & \rule{0pt}{11pt}\cellcolor{mygray2}Fine-tuned & \rule{0pt}{11pt}Twin & \rule{0pt}{11pt}FREE & \rule{0pt}{11pt}\cellcolor{mypink}\textbf{DiDi-Merg.-L} \\ \midrule

    \cellcolor{mygray}& \cellcolor{mygray}MMLU-Pro & \cellcolor{mygray2}49.7 & \cellcolor{mygray2}51.3 & 50.2 & \underline{50.4} & \cellcolor{mypink}\textbf{51.1} & \cellcolor{mygray2}54.0 & \cellcolor{mygray2}55.6 & 54.5 & \underline{54.8} & \cellcolor{mypink}\textbf{55.2}  \\

    \cellcolor{mygray}\textbf{General} & \cellcolor{mygray}MMLU-redux & \cellcolor{mygray2}70.5 & \cellcolor{mygray2}73.0 & \underline{72.6} & 72.4 & \cellcolor{mypink}\textbf{72.8} & \cellcolor{mygray2}75.1 & \cellcolor{mygray2}76.6 & 74.8 & \underline{75.2} & \cellcolor{mypink}\textbf{76.2} \\

    \cellcolor{mygray}& \cellcolor{mygray}GPQA-Diamond & \cellcolor{mygray2}33.6 & \cellcolor{mygray2}37.7 & \underline{36.1} & 35.9 & \cellcolor{mypink}\textbf{37.2} & \cellcolor{mygray2}34.7 & \cellcolor{mygray2}38.1 & 36.8 & \underline{37.2} & \cellcolor{mypink}\textbf{37.4} \\
    \midrule

    \multirow{3}{*}{\cellcolor{mygray}\textbf{Mathematics}} & \cellcolor{mygray}GSM8K~$_\text{(0 shot, CoT)}$ & \cellcolor{mygray2}85.9 & \cellcolor{mygray2}88.8 & 86.4 & \textbf{88.2} & \cellcolor{mypink}\underline{88.0} & \cellcolor{mygray2}91.7 & \cellcolor{mygray2}92.6 & 91.3 & \underline{91.3} & \cellcolor{mypink}\textbf{91.6} \\

    \cellcolor{mygray}& \cellcolor{mygray}MATH~$_\text{(0 shot, CoT)}$ & \cellcolor{mygray2}50.7 & \cellcolor{mygray2}56.2 & \underline{55.2} & 55.0 & \cellcolor{mypink}\textbf{55.3} & \cellcolor{mygray2}75.0 & \cellcolor{mygray2}75.3 & 72.1 & \textbf{74.2} & \cellcolor{mypink}\underline{74.0} \\

    \cellcolor{mygray}& \cellcolor{mygray}AMC 23~$_\text{(0 shot, CoT)}$ & \cellcolor{mygray2}25.0 & \cellcolor{mygray2}37.5 & 27.5 & \underline{27.5} & \cellcolor{mypink}\textbf{35.0} & \cellcolor{mygray2}52.5 & \cellcolor{mygray2}57.5 & 50.0 & \underline{47.5} & \cellcolor{mypink}\textbf{52.5} \\
    \midrule
    \multirow{2}{*}{\cellcolor{mygray}\textbf{Coding}} & \cellcolor{mygray}HumanEval~$_\text{(0 shot)}$ & \cellcolor{mygray2}68.3 & \cellcolor{mygray2}72.0 & 68.8 & \underline{69.1} & \cellcolor{mypink}\textbf{70.4} & \cellcolor{mygray2}85.4 & \cellcolor{mygray2}85.7 & 81.9 & \underline{84.1} & \cellcolor{mypink}\textbf{84.4} \\

    \cellcolor{mygray}& \cellcolor{mygray}MBPP~$_\text{(0 shot)}$ & \cellcolor{mygray2}66.9 & \cellcolor{mygray2}73.0 & 70.3 & \underline{71.2} & \cellcolor{mypink}\textbf{71.8} & \cellcolor{mygray2}80.2 & \cellcolor{mygray2}84.8 & 81.6 & \underline{82.5} & \cellcolor{mypink}\textbf{83.3} \\
    \midrule
    \cellcolor{mygray}\textbf{Instruction} & \cellcolor{mygray}AlpacaEval-2~$_\text{(LC \%)}$ & \cellcolor{mygray2}28.3 & \cellcolor{mygray2}65.4 & 52.4 & \underline{54.1} & \cellcolor{mypink}\textbf{59.5} & \cellcolor{mygray2}34.2 & \cellcolor{mygray2}63.6 & 53.3 & \underline{53.6} & \cellcolor{mypink}\textbf{58.7} \\

    \cellcolor{mygray}\textbf{Following} & \cellcolor{mygray}MT-Bench & \cellcolor{mygray2}8.4 & \cellcolor{mygray2}9.0 & 8.2 & \underline{8.3} & \cellcolor{mypink}\textbf{8.7} & \cellcolor{mygray2}8.4 & \cellcolor{mygray2}9.0 & 7.8 & \underline{8.2} & \cellcolor{mypink}\textbf{8.7} \\

    \midrule
    \multicolumn{2}{c}{\cellcolor{mygray}\textbf{Average}} & \cellcolor{mygray2}48.7 & \cellcolor{mygray2}56.4 & 52.8 & \underline{53.2} & \cellcolor{mypink}\textbf{55.0} & \cellcolor{mygray2}59.0 & \cellcolor{mygray2}63.8 & 60.4 & \underline{60.8} & \cellcolor{mypink}\textbf{62.2} \\

    \midrule
    \multicolumn{2}{c}{\cellcolor{mygray}\textbf{Parameters}} & \cellcolor{mygray2} x1 & \cellcolor{mygray2}x4 & x2.25 & x2.08 & \cellcolor{mypink}\textbf{x2.0} & \cellcolor{mygray2}x1 & \cellcolor{mygray2}x4 & x2.25 & x2.08 & \cellcolor{mypink}\textbf{x2.0}  \\
\bottomrule
\end{NiceTabular}}
\end{table}

%% file: tabs/mllm.tex
\begin{table}[!t]
\renewcommand{\arraystretch}{1.4}
\centering
\caption{Results of multi-modality dynamic merging experiments. The
performance retention is shown in parentheses.}
\label{tabs:mllm_result}
\resizebox{\linewidth}{!}{
\begin{tabular}{c|c|cccc}
\bottomrule
\rowcolor{mygray}
Task ($\rightarrow$) &  & {\textbf{2 Point Tasks}}   &  \textbf{3 Audio Tasks} &  \textbf{2 Video Tasks} & {\textbf{7 Image Tasks}} \\

\rowcolor{mygray}
Method ($\downarrow$) & \multirow{-2}{*}{\textbf{$\#$Params\textcolor{darkred}{($\downarrow$)}}} & Score (\%) / Acc. (\%) & Score (\%) / Acc. (\%) & Acc. (\%) & Acc. (\%) \\
\hline
\rowcolor{mygray2}{Individuals} &  [x2, x7]  &23.15 / 21.27 & 25.30 / 24.71& 39.79 & 62.23 \\
{NaiveMC}\pub{ACL2024}~\citep{18} &  [x1.24, x1.84] & \foo{22.65}{97.8} / \foo{20.49}{96.3} &\foo{24.59}{97.2} / \foo{30.65}{124.8}  & \foo{36.92}{93.0} &\foo{52.56}{83.6} \\

{MMER}\pub{ACL2025}~\citep{mmer} &  [x2.13, x2.44] & \underline{\foo{23.14}{99.9} / \foo{22.49}{105.7}} & \foo{25.20}{99.6} / \foo{38.51}{155.6}
&\underline{\foo{39.28}{98.5}}
& \foo{62.40}{100.3} \\

{FREE-Merging}\pub{ICCV2025}~\citep{free}  & [x2.04, x2.14] & \foo{23.02}{99.5} / \foo{21.14}{99.4} & \underline{\foo{25.23}{99.7} / \foo{38.58}{156.1}} & \foo{38.52}{96.8} & \underline{\foo{62.54}{100.5}} \\

\rowcolor{mypink}\textbf{\ourapproach (ours)} & [x1.4, x2.0]  & \textbf{\foo{23.08}{99.7}} / \textbf{\foo{22.58}{106.2}} & {\textbf{\foo{25.27}{99.9} / \textbf{\foo{38.74}{156.8}}}} & \textbf{\foo{39.51}{99.3}} & \textbf{\foo{62.93}{101.1}} \\
\hline
\end{tabular}
}
\end{table}

%% file: sec/6_analysis.tex
\input{tabs/ablation_time_side}

\section{Analysis}
\label{sec:analysis}
We evaluate component contributions and solving cost. Additional rank,
PEFT, and task-relationship analyses are reported in the appendix.

\subsection{Ablation Study}
\label{sec:ablation}
Tab.~\ref{tab:ablation} ablates the design
choices of \ourapproach\ at a matched parameter budget. Removing the
shared low-rank matrix or the per-expert optimal rank degrades
ViT-B/32 accuracy by $2.5\%$ and $2.4\%$, respectively, while
disabling layer-wise rank optimization causes a smaller drop.
Removing low-rank refinement reduces ViT-B/32 accuracy from $89.8$ to
$85.5\%$ ($4.3$ points) and ViT-L/14 from $95.6$ to $91.2\%$
($4.4$ points), showing that rank selection alone does not account for
the complete method. SVD-based initialization of the LoRA
factors has a small effect on the final score ($-0.2\%$) but
substantially accelerates convergence.

\paragraph{Is the rank optimization necessary?}
A static merge followed by uniform-rank recovery is a simpler alternative
to learned allocation.
App.~\ref{app:rank_alloc_control} tests this with four static mergers
(Simple Averaging, TIES-Merging, Iso-CTS, WUDI-Merging) at a matched
$1.24\times$ budget. The four alternatives reach $80.4$, $81.5$, $82.2$, and $84.6\%$,
compared with $89.3\%$ for \ourapproach-S: gaps of $4.7$--$8.9$
percentage points. This comparison supports the complete adaptive
allocation design; a fully crossed allocation-by-refinement experiment
would provide a more specific interaction analysis.

\subsection{Computation Resource}
\label{sec:compute}
Tab.~\ref{tab:time} compares solving time and GPU memory on
ViT-B/32 against optimisation-based merging baselines. \ourapproach\
uses $14$ min $37$ s in total, compared with $3$ min $27$ s for
TSV-C and $1$ min $54$ s for WUDI in this timing table. Its memory use
is $1.1$\,GB, below WUDI's $4.0$\,GB; the lower GPU memory
requirement therefore does not imply lower wall-clock cost. The Find-Optimal-Rank stage runs sequentially per task
and is the dominant wall-clock bottleneck of the pipeline.

\paragraph{Storage accounting.}
All total-storage ratios count one backbone and the shared and expert
factors. The task-ID selector introduces no learned router parameters.
Appendix~\ref{app:storage} gives the bit-level decomposition and learned
rank statistics, including the global treatment of masks and scalars.

%% file: tabs/ablation_time_side.tex
\begin{figure}[!t]
\centering
\begin{minipage}[t]{0.50\linewidth}
\vspace{0pt}
  \centering
  \captionof{table}{Component ablation on \ourapproach\ at a matched
  parameter budget. When an optimal-rank component is ablated, its
  rank is replaced with a budget-preserving uniform value.}
  \label{tab:ablation}
  \resizebox{\linewidth}{!}{
  \begin{tabular}{lcc}
  \toprule
  \rowcolor{mygray}\textbf{Method} & \textbf{ViT-B/32} & \textbf{ViT-L/14} \\
  \midrule
  \rowcolor{mypink}\textbf{\ourapproach\ (full)} & \textbf{89.8} & \textbf{95.6} \\
  \midrule
  \; $-$ w/o Shared Matrix                       & 87.3 & 93.7 \\
  \; $-$ w/o Shared Optimal Rank                 & 88.9 & 94.7 \\
  \; $-$ w/o Expert Optimal Rank                 & 87.4 & 93.6 \\
  \; $-$ w/o Layer-wise Optimal Rank             & 88.2 & 94.1 \\
  \rowcolor{mylightblue}\; $-$ w/o Low-rank Module Update & 85.5 & 91.2 \\
  \; $-$ w/o Low-rank Module Init.               & 89.6 & 95.5 \\
  \bottomrule
  \end{tabular}
  }
\end{minipage}\hfill
\begin{minipage}[t]{0.48\linewidth}
\vspace{0pt}
  \centering
  \captionof{table}{Solving time and GPU memory on ViT-B/32, against
  optimisation-based merging baselines. Solving time and memory describe different resource costs.}
  \label{tab:time}
  \renewcommand\arraystretch{1.05}
  \setlength{\tabcolsep}{4pt}
  \resizebox{\linewidth}{!}{
  \begin{tabular}{l|ccc}
  \toprule
  \rowcolor{mygray}\textbf{Method} & \textbf{Acc.} & \textbf{Time} & \textbf{GPU Mem.} \\
  \midrule
  TIES Merging\pub{NeurIPS23}~\cite{ties}      & 72.4 & 4\,s          & 0\,GB     \\
  Adamerging\pub{ICLR24}~\cite{adamerging}      & 81.1 & 127\,min      & 17.1\,GB  \\
  TSV-C\pub{CVPR25}~\cite{tsv}                  & 89.1 & 3\,min\,27\,s & 0\,GB     \\
  WUDI-Merging\pub{ICML25}~\cite{wudi}          & 85.2 & 1\,min\,54\,s & 4.0\,GB   \\
  \rowcolor{mypink}\textbf{\ourapproach\ (Ours)} & $\mathbf{89.8}$ & 14\,min\,37\,s & 1.1\,GB \\
  \rowcolor{mypink}\; $-$Find Optimal Rank       & --   & 10\,min\,22\,s & 1.1\,GB  \\
  \rowcolor{mypink}\; $-$Low-rank Module Update  & --   & 4\,min\,15\,s  & 0.6\,GB  \\
  \bottomrule
  \end{tabular}
  }
\end{minipage}
\end{figure}

%% file: sec/7_conclusion.tex
\section{Conclusions}
\label{sec:conclusion}
DiDi-Merging consolidates fine-tuned models through a shared--expert
low-rank decomposition, coupled rank allocation, and data-free refinement.
On the evaluated eight-task vision benchmark, it reaches $89.3\%$ at
$1.24\times$ storage and $89.8\%$ at $1.4\times$, including the backbone.
Language and multimodal results extend the evaluation at their own
budgets. The present evidence supports this accuracy--storage trade-off
under known task identity; practical routing and task-growth behavior
remain separate empirical questions.

%% file: camera_ready_disclosures.tex
\begin{ack}
\textbf{Funding statement.}
This work was supported by the Hong Kong Polytechnic University Internal Research Fund 
(the Projects of RIAIoT, Project ID P0065264), UDomain Web Hosting Company Limited (Collaboration Scheme, Project ID P0064507), and the Innovation and Technology Fund--Guangdong-Hong Kong Technology Cooperation Funding Scheme (ITF-TCFS) (Project ID P0049179).

\textbf{Competing interests.}
The authors declare that they have no competing interests.
\end{ack}

%% file: sec/8_appendix.tex

\section{Novelty and Contribution}
\label{app:a}
DiDi-Merging combines a shared--expert decomposition, differentiable
rank allocation under a coupled budget, and parameter-space refinement.
Its contribution is the consolidation formulation and evaluated
accuracy--storage trade-off, rather than a new SVD or low-rank primitive.

Static methods~\citep{ties,pcb} form one input-agnostic policy. In contrast,
DiDi retains task-specific capacity selected by known task identity in the
main evaluation. TALL-Mask and EMR~\citep{talls,emr} represent task
variation using masks and scalars around shared dense state; Twin and
FREE~\citep{twin,free} use their corresponding shared and expert
representations. DiDi compresses both shared and expert matrices through
joint rank allocation. Storage comparisons must use a common convention
for the backbone, auxiliary state, and any learned router.

TSV-C and SMILE~\citep{tsv,smile} provide other expert-based operating
points. Comparisons with these methods in the result tables concern the
specified checkpoints and budgets. LoRA-MoE~\citep{zhang2023composing}
addresses mixtures of low-rank modules and routing; DiDi's fitting signal
instead comes from already fine-tuned model parameters. Retaining
separate task components makes conditional reconstruction possible, but
does not establish forget-free incremental learning or constant accuracy
as tasks are added under a fixed total budget.

For broader multimodal context, related studies examine hierarchical
shared--private representation learning, governed agent collaboration,
reinforcement-learned panoramic scanpaths for quality assessment,
and temporal answer grounding in instructional video and multimodal STEM
reasoning evaluation~\citep{meng2026ahra,meng2026gcl,wang2026rlscaniqa,qi2026cacr,
jin2026unveilingfinegrainedvisualtraces}.
Complementary visual adaptation work studies instance-aware prompting,
cascaded semantics, layer-specific prompt fusion, structured low-rank updates,
and counterfactual visual alignment~\citep{xiao2025visual,xiao2026adapting,
xiao2026layer,xiao2026not,xiao2026staying}. Earlier visual studies also
investigate evolutionary refinement of pretrained networks and RAW-image
restoration~\citep{du2024impacts,du2023end}.

\section{Additional Analysis}
\label{app:b}

\subsection{Effectiveness of Differentiable Rank Allocation}
\label{app:b1}
To further demonstrate the effectiveness of our differentiable rank
allocation strategy, we compared it with several non-optimization-based
rank allocation methods. In the ViT-B/32 vision experiments, we
constrained the total rank budget to $0.4\times$ the number of parameters
and evaluated uniform allocation, random allocation, and energy-based
allocation. The energy-based method dynamically assigns ranks by
computing the cumulative singular value energy across different tasks
and layers. For random allocation, we report the average results over
five runs. As shown in Tab.~\ref{tab:rank_effect}, random allocation can
degrade model performance, while the energy-based strategy offers
minimal improvement, underscoring the advantage of our approach.
\input{tabs/rank_effect}

\subsection{Controlled Comparison Against Static Merge \texorpdfstring{$+$}{+} Uniform-rank LoRA}
\label{app:rank_alloc_control}
To address the concern that the gains of \ourapproach\ might be
attributed primarily to the LoRA recovery stage rather than to
differentiable rank allocation, we constructed a controlled comparison
at a strictly matched parameter budget. We pair four static merging
strategies (Simple Averaging, TIES-Merging, Iso-CTS, and
WUDI-Merging) with a LoRA recovery stage whose rank is fixed to a
single uniform value $r$ across all tasks and layers. The rank $r$ is
selected so that the total parameter count (backbone $+$ shared $+$
per-task LoRAs) falls within $\pm 2\%$ of $1.24\times$ the single
fine-tuned model, matching the budget of \ourapproach-S.

Table~\ref{tab:rank_alloc_control} reports the results. The WUDI plus
uniform-recovery configuration is a relevant alternative, but the table
compares complete configurations rather than a fully crossed factorial.
The four controls reach $80.4$--$84.6\%$, compared with $89.3\%$ for
DiDi-S. This supports the evaluated adaptive design at similar storage;
a separate allocation-by-refinement factorial is needed to quantify their
interaction or assign the complete gap to one component.
\input{tabs/ablation_baselines}

\subsection{Hyperparameter Analysis}
\label{app:b2}
Tab.~\ref{tab:hyperparams} summarizes the hyperparameters used for the
different models. While the optimal settings vary slightly depending on
the model type, \ourapproach remains stable when these values fluctuate
within a reasonable range. Sensitivity analyses on key hyperparameters
($\beta$, $\gamma$, and learning rates) yield consistent results,
suggesting that our method is largely insensitive to specific
configurations. Moreover, varying the static merging strategy used to
initialize the shared expert consistently yields comparable results,
further demonstrating the robustness of our approach.

We further analyzed the accuracy of dynamic model merging across
different models, including vision models (ViT-B/32), large language
models (LLaMA3-8B), and PEFT models (QWEN-14B finetuned with LoRA), as
shown in Tab.~\ref{tab:hyper_app}. Among all the hyperparameters,
$\beta$ and the learning rates mainly affect optimization speed, with
very little impact on accuracy. Different choices of loss functions and
$\gamma$ can have some effect. Specifically, $\gamma$ is introduced to
normalize the task reconstruction loss across different models, since
task vectors vary in magnitude across tasks, different normalization
values are required to keep the loss comparable to the compression
ratio.
\input{tabs/hyper_combined}

\subsection{Performance Gain Beyond Individual Modalities in MLLM Merging}
\label{app:b3}
In dynamic model fusion for MLLMs, it has been observed that the merged
models can achieve performance exceeding that of any single individual
modality, sometimes reaching more than $100\%$ relative to the original
single-modality performance. Our MLLM dynamic merging experiments
indicate that this improvement is not due to simply removing redundant
parameters, but rather results from the shared knowledge across
different modalities. This shared knowledge provides additional
information, such as prior knowledge or instruction-following
capabilities, which enhances model performance. When the shared
knowledge is removed, the performance gain disappears, confirming its
crucial role. These analyses agree with prior findings~\cite{mmer}.

\subsection{Comparison of PEFT Merging Strategies}
\label{app:peft_comparison}
We compare \ourapproach\ against representative PEFT merging
strategies along several practical dimensions
(Tab.~\ref{tab:peft_comparison}). KnOTS~\cite{knots} and Core Space
are static methods that produce a full matrix or a shared basis
without routing, leading to performance degradation when many tasks
must coexist. Twin-Merging~\cite{twin} and FREE-Merging~\cite{free}
introduce a router and recover near-lossless performance, but at
the cost of heavy per-task parameter overhead. \ourapproach\
preserves the dynamic-routing benefits of the latter group while
keeping the per-task expert footprint slim through differentiable
rank allocation.
\input{tabs/peft_comparison}

\paragraph{Implementation on LoRA modules.}
\ourapproach\ extends seamlessly to PEFT settings, where each
fine-tuned model is parameterised as
$\boldsymbol{\theta}_t = \boldsymbol{\theta}_0 + \mathrm{LoRA}_t$
with $\mathrm{LoRA}_t$ a low-rank module specific to task $t$. The
task vector reduces to the LoRA module itself,
$\boldsymbol{\tau}_t = \boldsymbol{\theta}_t - \boldsymbol{\theta}_0
= \mathrm{LoRA}_t$, so the static merge of task vectors is
$\boldsymbol{M}_s = \mathrm{avg}(\boldsymbol{\tau}_1, \ldots,
\boldsymbol{\tau}_T) = \mathrm{LoRA}_s$, and each expert residual is
$\boldsymbol{M}_t = \boldsymbol{\tau}_t - \boldsymbol{M}_s
= \mathrm{LoRA}_t - \mathrm{LoRA}_s$. The dynamic merging form
(Eq.~\ref{eq:dynmerge}) then specialises to:
\begin{equation}
\begin{aligned}
\boldsymbol{\theta}^{*}(\boldsymbol{x})
&= \boldsymbol{\theta}_0 + \tilde{\boldsymbol{M}}_s
   + \sum_{t=1}^{T} w_t(\boldsymbol{x})\,\tilde{\boldsymbol{M}}_t \\
&= \boldsymbol{\theta}_0 + \widetilde{\mathrm{LoRA}_s}
   + \sum_{t=1}^{T} w_t(\boldsymbol{x})\,
     \widetilde{\mathrm{LoRA}_t - \mathrm{LoRA}_s} \\
&= \boldsymbol{\theta}_0 + \mathrm{LoRA}^{*},
\end{aligned}
\end{equation}
where $\widetilde{\cdot}$ denotes the differentiable rank-truncated
version of a matrix and $\mathrm{LoRA}^{*} =
\widetilde{\mathrm{LoRA}_s} + \sum_{t=1}^{T} w_t(\boldsymbol{x})\,
\widetilde{\mathrm{LoRA}_t - \mathrm{LoRA}_s}$ is the merged LoRA
module. Hence \ourapproach\ on the full model
$\boldsymbol{\theta}$ is equivalent to \ourapproach\ applied directly
to the LoRA factors: the backbone $\boldsymbol{\theta}_0$ never
enters the optimisation, and only the LoRA factors are stored,
decomposed, and refined. This makes the framework directly
compatible with existing PEFT pipelines. Empirical results validating
this equivalence on Qwen-14B LoRA adapters are reported in
App.~\ref{app:peft_results}.

\subsection{Task Relationships and Rank Allocation}
\label{sec:task_relat}
\begin{figure}[!t]
\centering
\includegraphics[width=0.65\linewidth]{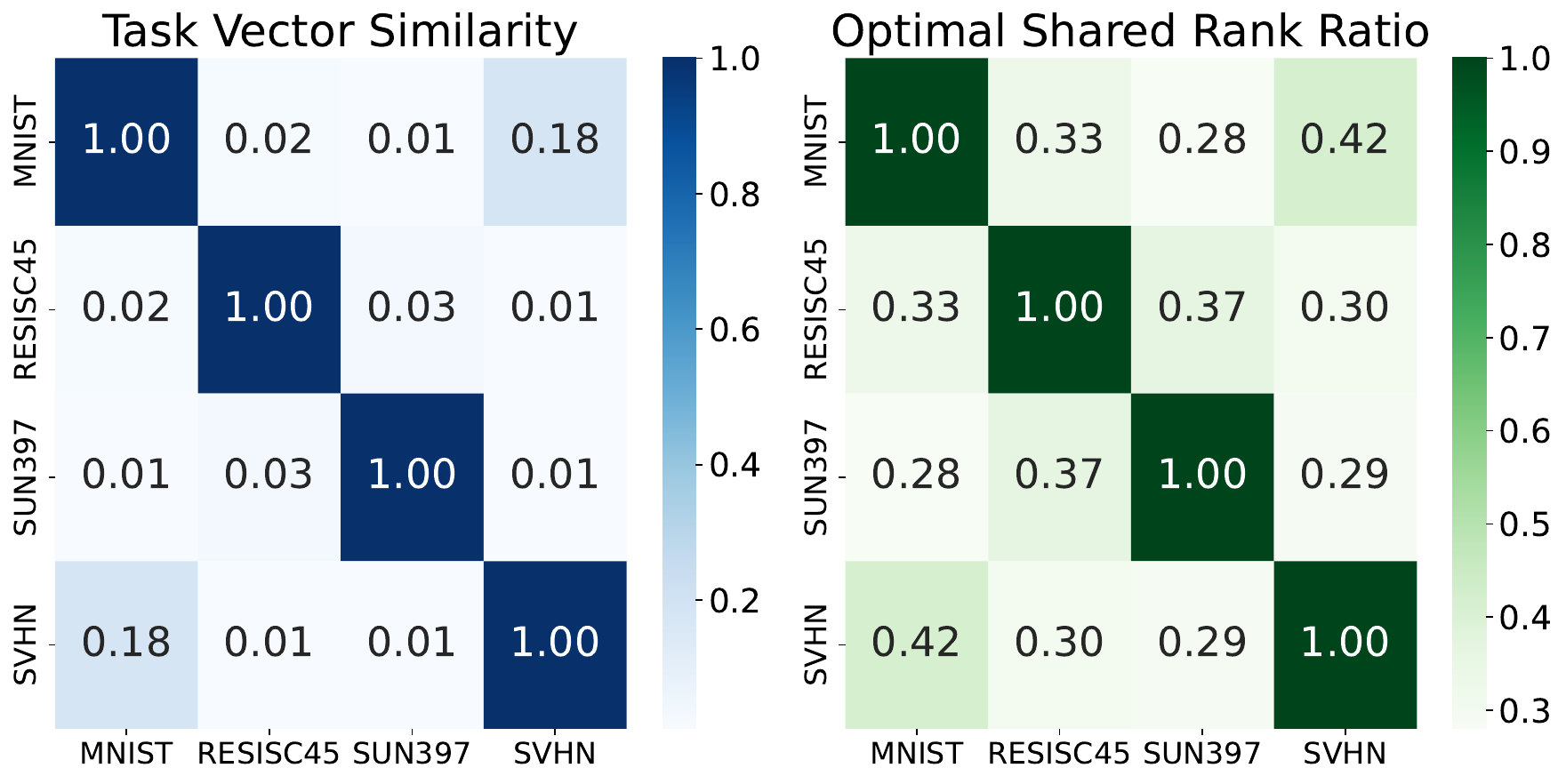}
\caption{Task-vector similarity (left) vs.\ optimal shared-rank
ratio (right) for pairs of four tasks.}
\label{fig:heatmap}
\end{figure}
Fig.~\ref{fig:heatmap} compares pairwise task-vector similarity with
the optimal shared-rank ratio across four vision tasks, and the two
heatmaps align tightly. MNIST and SVHN, both digit-classification
tasks, form the most similar pair (similarity $0.18$) and receive the
largest shared-rank ratio ($0.42$), while less related pairs settle
at a lower $0.28$. This mapping emerges purely from optimisation,
without hand-tuned similarity priors, and is consistent with the separately reported extension from 8
to 20 vision tasks (Tab.~\ref{tab:vision2}) retains $\sim\!99\%$ of
fine-tuned accuracy without re-tuning the rank budget.

\subsection{Storage Accounting}\label{app:storage}
\label{sec:storage}
Table~\ref{tab:storage_bit} gives the per-module bit cost across
methods. Heavy-shared baselines duplicate the backbone in their
shared component ($32mn$); heavy-expert baselines pay
$32T r_t(m{+}n)$ on uncompressed experts. Differentiable allocation
keeps both small ($\bar{r}_s, \bar{r}_t \!\sim\! 15$ on ViT-B/32;
App.~\ref{app:storage}), giving the $\le\!1.4\times$ total budget.
The optimised ranks themselves reflect architectural structure:
attention-output projections receive the smallest rank, MLP modules
the largest, and the ordering is preserved across the $1.24\times$
and $1.4\times$ budgets, matching the relative information density
of each module type.
\input{tabs/storage_bit}

This section makes the parameter budgets reported in the main paper
($1.24\times / 1.4\times / 2.0\times$) explicit and clarifies how
they are computed. Let $P_{\rm model}$ denote all parameters in one pretrained backbone,
$R$ the number of parameters in a learned router, and $B_{\rm factors}$
the shared and task-specific factor count in Eq.~\ref{eq:coupled_budget}.
Total fp32 storage is
\begin{equation}
 32\big(P_{\rm model}+B_{\rm factors}+R\big)\quad\text{bits},
\end{equation}
and the normalized ratio is
$1+B_{\rm factors}/P_{\rm model}+R/P_{\rm model}$.
The backbone is included once. The main experiments use an ideal task-ID
selector, for which $R=0$; a learned router would be counted once at
model level. Per-module tables decompose the weight terms, and their
router column must not be charged again for every layer.

The bit-level decomposition across baselines is reported in
Tab.~\ref{tab:storage_bit} (this appendix).

\paragraph{Average optimal ranks across modules.}
The ranks found by the differentiable optimisation vary across
module types (because $d_{\text{in}}, d_{\text{out}}$ differ) and
across tasks (because the spectra of task vectors differ).
Tab.~\ref{tab:storage_rank} reports the average optimal rank for
each module type in ViT-B/32, aggregated over the $12$ transformer
blocks and the $T{=}8$ vision tasks at the $1.24\times$
(\ourapproach-S) and $1.4\times$ (\ourapproach-M) budgets.
Two patterns emerge: (i) ranks scale with module size: attention
output ($768{\times}768$) requires the smallest rank, attention
input ($2304{\times}768$) is intermediate, and MLP modules
($3072{\times}768$ and $768{\times}3072$) require the largest;
and (ii) the per-module ratio between the two budgets is roughly
constant ($\sim\!1.7\times$), consistent with the linear scaling
of the budget delta from $0.24\times$ to $0.4\times$, indicating
that the optimisation rescales ranks proportionally rather than
redistributing them across modules.
\input{tabs/storage_rank}

A finer-grained breakdown that separates the shared and per-task
expert ranks is reported in Tab.~\ref{tab:storage_rank_detailed}.
Within each module type, $\bar{r}_s$ and $\bar{r}_t$ are tightly
correlated, indicating the shared module captures cross-task
structure that each task vector also relies on. Plugging the
finer-grained averages back into Eq.~\ref{eq:coupled_budget} recovers the $1.24\times$
budget within rounding noise.
\input{tabs/storage_rank_detailed}

\section{Additional Results}
\label{app:c}

\subsection{Detailed PEFT Merging Results}\label{app:peft_results}
PEFT methods~\cite{peft} efficiently adapt large language models to
new tasks, but storing separate adapters per task can be
costly~\cite{zhang2023composing}. We merge LoRA~\cite{hulora}
adapters of Qwen-14B~\cite{bai2023qwen} on three generative
tasks~\cite{mmlu}, following~\cite{twin,free}. As
Tab.~\ref{tab:peft_qwen_lora_all} shows, DiDi-Merging-M matches
prior baselines with fewer parameters, and DiDi-Merging-L matches
their parameter overhead while attaining $0.7\%$ higher performance.
\input{tabs/lora}

\subsection{Detailed Multimodal Results}
We report the detailed results of each method on the corresponding
datasets, including vision-language (Tab.~\ref{tab:mllm_I}), point-cloud
(Tab.~\ref{tab:mllm_P}), audio, and video (Tab.~\ref{tab:mllm_AV})
tasks, illustrating that DiDi-Merging consistently achieves superior
performance while maintaining a slim parameter footprint. As an
experimental detail, we evaluate DiDi-Merging on multiple multimodal
tasks, ensuring that all models are fine-tuned under identical
environments. Performance is measured in terms of accuracy or overall
scores, as well as the percentage of performance retained relative to
each model's fine-tuned counterpart. We compare DiDi-Merging with representative data-free dynamic mergers.
\input{tabs/mllm_I}
\input{tabs/mllm_P}
\input{tabs/mllm_AV}

\section{Method Details}
\label{app:g}

\subsection{Differentiable Optimization of Ranks}
\label{app:g1}
To efficiently allocate ranks for shared and expert modules, we adopt a
differentiable approach inspired by smooth SVD truncation. Each
parameter matrix is decomposed via SVD, and a smooth truncation function
with a learnable continuous rank $r$ produces a low-rank approximation
while keeping the rest of the module parameters frozen. The
reconstructed task vectors, obtained by combining truncated shared and
expert matrices, are optimized with a task reconstruction loss. A
compression-aware term penalizes deviations from a target rank budget,
balancing task fidelity and parameter efficiency. By jointly optimizing
$r$ across layers, the method adaptively allocates ranks between shared
and expert matrices to maintain compact yet informative low-rank
modules.

\subsection{Optimization for Low-rank Module Update}
During refinement, the shared and task-specific low-rank factors are
updated with fixed ranks, while all pretrained backbone parameters remain frozen. The goal of
this stage is to efficiently recover task-specific performance that may
have been lost during low-rank truncation. The optimization is
performed as follows:
\begin{itemize}
    \item \textbf{Optimizer:} Adam with a small learning rate, typically
    in the range $1\mathrm{e}{-4}$ to $1\mathrm{e}{-3}$.
    \item \textbf{Batch-free:} Optimization is fully data-free; the
    supervision signal comes solely from the stored task vectors
    $\boldsymbol{\tau}_t$, so no original training data are required.
    \item \textbf{Regularization:} Optionally, a term
    $\lambda \|\boldsymbol{A}_{t,l} \boldsymbol{B}_{t,l} -
    \boldsymbol{M}^{(0)}_{t,l}\|_F^2$ or standard weight decay can be
    applied to stabilize updates.
    \item \textbf{Multi-task Optimization:} When optimizing multiple
    tasks simultaneously, different task losses are balanced using
    dynamic weighting. A temperature-based scaling factor is applied to
    adjust the relative contribution of each loss over time, ensuring
    that no single task dominates the optimization and that all tasks
    maintain stable improvement.
\end{itemize}
This lightweight procedure produces compact per-task low-rank modules
that respect the allocated rank budget, restoring task-specific
fidelity while maintaining high efficiency and performance.

\input{sec/7_limitations}

\section{Notation}\label{app:notation}
Tab.~\ref{tab:notation} summarises the key symbols used throughout
the paper.

\begin{table}[!t]
\centering
\caption{Notation summary.}
\label{tab:notation}
\renewcommand\arraystretch{1.05}
\begin{tabular}{ll}
\toprule
\rowcolor{mygray}\textbf{Symbol} & \textbf{Meaning} \\
\midrule
\multicolumn{2}{l}{\emph{Problem setting}} \\
$T$ & Number of tasks \\
$L$ & Number of layers \\
$\boldsymbol{\theta}_0$ & Pre-trained backbone parameters \\
$\boldsymbol{\theta}_t$ & Fine-tuned parameters for task $t$ \\
$\boldsymbol{\tau}_t = \boldsymbol{\theta}_t - \boldsymbol{\theta}_0$ & Task vector for task $t$ \\
$\lambda_t$ & Coefficient of $\boldsymbol{\tau}_t$ in static merging \\
\midrule
\multicolumn{2}{l}{\emph{Dynamic merging form (Eq.~\ref{eq:dynmerge})}} \\
$\boldsymbol{M}_s$ & Shared module \\
$\boldsymbol{M}_t$ & Per-task expert module \\
$w_t(\boldsymbol{x})$ & Input-dependent routing weight for task $t$ \\
\midrule
\multicolumn{2}{l}{\emph{Low-rank decomposition (\S\ref{subsec:method2})}} \\
$d_{\text{in}}, d_{\text{out}}$ (or $m_l, n_l$) & Module weight dimensions \\
$r_s, r_t$ & Shared, per-task expert ranks \\
$r_{s,l}, r_{t,l}$ & Per-layer shared / per-task ranks at layer $l$ \\
$\boldsymbol{U}, \boldsymbol{\Sigma}, \boldsymbol{V}$ & SVD factors \\
$\sigma_i$, $\tilde{\sigma}_i$ & Singular values; smooth-truncated values \\
$\beta$ & Smoothness factor of the truncation function \\
\midrule
\multicolumn{2}{l}{\emph{Low-rank module update (\S\ref{subsec:method3})}} \\
$\boldsymbol{A}_{t,l}, \boldsymbol{B}_{t,l}$ & LoRA factor matrices for task $t$ at layer $l$ \\
$\boldsymbol{A}_{s,l}, \boldsymbol{B}_{s,l}$ & Shared LoRA factor matrices at layer $l$ \\
$L_{\text{task}}$ & Task reconstruction loss \\
$R_{\text{now}}, R_{\text{tar}}$ & Current / target compression ratio \\
$\gamma$ & Weight for compression-ratio penalty \\
\midrule
\multicolumn{2}{l}{\emph{Storage accounting (App.~\ref{app:storage})}} \\
$P = d_{\text{in}} d_{\text{out}}$ & Per-module parameter count \\
$R$ & Router parameter count \\
\bottomrule
\end{tabular}
\end{table}

\section{Broader Impact}\label{app:broader_impact}
\ourapproach\ reduces the storage and serving cost of multi-task
systems, lowering the deployment barrier for collections of
specialised models. Like all efficient deployment methods, lower
cost can amplify both beneficial uses (low-resource personalisation,
on-device multi-task inference) and harmful ones (easier serving of
biased or unsafe fine-tunes). \ourapproach\ inherits the biases of
its source checkpoints; we introduce no additional training data
or labels. We see particular promise in high-stakes multimodal
deployments such as medical diagnosis, where heterogeneous expert
models for imaging, language, and clinical reasoning must coexist
under tight storage and latency budgets~\cite{zhu2025pathology,
zhu2026medeyes, lin2026medcausalx,
zhu2026medsynapsevbridgingvisualperception,
wei2026open, chen2026vocabulary}; slim dynamic merging offers a
path to deploy such expert ensembles without retraining or exposing
the underlying patient data.

\section{Baselines details}
\label{app:d}
This section provides a detailed description of the baselines, as
outlined below.
\begin{itemize}
    \item \textbf{Task Arithmetic}~\citep{ta} first defines the concept
    of ``task vectors'' and merges these vectors into a pre-trained model
    to execute multi-task learning. The model is produced by scaling and
    adding the task vectors to the initial model as
    $\theta_m = \theta_\textrm{init} + \lambda \cdot \sum_{t=1}^n \tau_t$.
    \item \textbf{Ties-Merging}~\citep{ties} further solves the task
    conflict problem in Task Arithmetic~\citep{ta}. It eliminates
    redundant parameters and resolves symbol conflicts through three
    steps: Trim, Elect Sign, and Disjoint Merge.
    \item \textbf{DARE}~\citep{dare} sets the majority of delta
    parameters to zero and rescales the rest by
    $\theta' = \theta \cdot (1/(1-p))$ where $p$ is the proportion of
    delta parameters dropped, therefore efficiently reducing parameter
    redundancy.
    \item \textbf{PCB-Merging}~\citep{pcb} effectively adjusts parameter
    coefficients through balancing parameter competition within model
    population.
    \item \textbf{CAT-Merging}~\citep{cat} introduces a training-free
    framework that trims conflict-prone task vector components via
    projection for linear weights and masking for normalization
    parameters.
    \item \textbf{TALL-Mask}~\citep{talls} localizes the task-specific
    information in a multi-task vector, which deactivates irrelevant
    parts for each task in the merged vector using binary task masks.
    \item \textbf{EMR-Merging}~\citep{emr} first selects a unified model
    from all weights, then generates lightweight task-specific
    modulators (masks and rescalers) to align direction and magnitude
    with each source model.
    \item \textbf{Twin-Merging}~\citep{twin} proposes a method that
    modularizes knowledge into shared and exclusive components, while
    compression reduces redundancy and improves parameter efficiency.
    \item \textbf{FREE-Merging}~\citep{free} reveals that task
    interference is evident in the frequency domain of model parameters.
    \item \textbf{TSV-Compression}~\citep{tsv} leverages low-rank space
    to define a new measure of task interference based on the interaction
    of singular vectors from different tasks.
    \item \textbf{NaiveMC}~\citep{18} composes existing MLLMs into a
    new model that retains each source model's modality-specific
    understanding.
    \item \textbf{MMER}~\citep{mmer} introduces a training-free approach
    for seamless multimodal expansion of LLMs through parameter merging
    and decoupling.
\end{itemize}

\section{Datasets details}
\label{app:e}
\textbf{SUN397}~\citep{xiao2016sun} is a scene classification dataset,
which contains images in 397 classes, with a total of 108{,}754 images,
and each class has at least 100 images.

\textbf{Stanford Cars (Cars)}~\citep{krause20133d} is a car
classification dataset, which contains 196 classes of cars and a total
of 16{,}185 images. Each class is split equally between the training
and test sets.

\textbf{RESISC45}~\citep{cheng2017remote} is a remote sensing image
scene classification dataset. It contains 45 classes of scenes and a
total of 31{,}500 images, with approximately 700 per class.

\textbf{EuroSAT}~\citep{helber2019eurosat} is a satellite image
classification dataset containing 27{,}000 labeled and geo-referenced
images in 10 classes.

\textbf{SVHN}~\citep{yuval2011reading} is a real-world digital
classification dataset extracted from house numbers in Google Street
View images. There are 10 classes in total. The training set contains
73{,}257 samples, the test set contains 26{,}032 samples, and 531{,}131
additional simple samples can be used as additional training data.

\textbf{GTSRB}~\citep{stallkamp2011german} is a traffic sign
classification dataset, which contains 43 classes of traffic signs with
a total sample size of more than 50{,}000.

\textbf{MNIST}~\citep{lecun1998mnist} is a benchmark dataset for image
classification. It contains grayscale images of handwritten digits in
10 classes. The number of images in the training and test sets is
60{,}000 and 10{,}000 respectively. The number of images in each class
is balanced.

\textbf{DTD}~\citep{cimpoi2014describing} is a texture classification
dataset, which contains 47 classes, a total of 5{,}640 images, with
each class having approximately 120 images.

\textbf{AlpacaEval-2}~\citep{AlpacaEval} comprises 805 instructions from
five different datasets and assesses models using two metrics:
length-controlled (LC) win rate and raw win rate
(WR)~\citep{dubois2024length}. GPT-4-Preview-1106 serves as both the
baseline model and the evaluator for the other models.

\textbf{MT-Bench}~\citep{zheng2023judging} contains 80 multi-turn
dialogues across eight categories, including writing, roleplay,
reasoning, math, coding, extraction, STEM, and humanities. Each
response is evaluated by GPT-4 on a scale from 1 to 10, with the
average score reported for each dialogue turn across the 80 dialogues.
We use GPT-4-0613 as the judge model following the official setting.

\textbf{MMLU-Pro}~\citep{wang2024mmlupro} is an enhanced version of the
MMLU dataset, designed to address issues such as noisy data and reduced
difficulty due to advances in model capabilities and increased data
contamination. MMLU-Pro increases challenge levels by expanding
multiple-choice options from 4 to 10, requiring reasoning across more
questions, and incorporating expert-reviewed annotations for improved
quality and reduced noise.

\textbf{MMLU-redux}~\citep{gema2024we} is a re-annotated subset of the
MMLU dataset created through manual assessment from 14 human experts.

\textbf{GPQA-Diamond}~\citep{rein2023gpqa} is a challenging knowledge
benchmark crafted by PhD-level domain experts in biology, physics, and
chemistry. The dataset contains questions that are straightforward for
experts but difficult for laypersons. We evaluate the highest quality
diamond set comprising 198 questions.

\textbf{Arena-Hard}~\citep{arenahard2024} is a challenging
instruction-following benchmark that closely aligns with the human
preference ranking from Chatbot Arena, a crowd-sourced platform for
evaluating LLMs. It spans 250 high-quality topic clusters including 500
well-defined technical problem-solving queries. We report the win rate
against GPT-4-0314 using GPT-4-Preview-1106 as the judge model.

\textbf{GSM8K}~\citep{cobbe2021gsm8k} contains grade-school math word
problems for evaluating mathematical reasoning.

\textbf{MATH}~\citep{hendrycks2021math} is a dataset of math problems
ranging in difficulty from middle school to high school competition
level. It tests a wide range of mathematical skills, including algebra,
calculus, number theory, and probability.

\textbf{AMC23}~\citep{yang2024qwen25math} refers to the 2023 American
Mathematics Competition, featuring 25 multiple-choice questions that
test advanced high school mathematics, including trigonometry, advanced
algebra, and elements of calculus.

\textbf{HumanEval}~\citep{chen2021evaluating} evaluates code generation
capabilities by presenting models with function signatures and
docstrings and requiring them to implement the function body in Python.

\textbf{MBPP}~\citep{austin2021program} is a dataset of simple
programming problems designed to assess the ability of models to
generate short Python code snippets from natural language descriptions.

\section{Implementation Details}
\label{app:f}
In this section, we provide detailed descriptions of the experimental
setup used for both LLM (\S\ref{app:f1}) and multimodal LLM
(MLLM)~(\S\ref{app:f2}) fine-tuning. We first outline the procedures for
obtaining task-specific fine-tuned LLMs and constructing the associated
datasets, followed by the fine-tuning protocol for MLLMs across multiple
modalities. All experiments are designed to ensure consistency and
comparability, with hyperparameters and training configurations aligned
with prior studies. All our experiments are conducted on an NVIDIA
$8{\times}$A800-SXM4-80GB machine.

\subsection{LLM Fine-tuning}
\label{app:f1}
In our LLM experiments, we follow a structured pipeline to fine-tune
one model for each expert task.

\noindent\textbf{SFT Experiments.} We use the \texttt{Llama-Factory}
library\footnote{\url{https://github.com/hiyouga/LLaMA-Factory}}~\citep{llamafactory}
for implementation. All target models are fine-tuned for 3 epochs with a
batch size of 128 and a maximum sequence length of 2048 tokens. A cosine
learning rate schedule with a warmup ratio of 0.1 is employed.

\noindent\textbf{DPO Experiments.} We use the
\texttt{alignment-handbook}\footnote{\url{https://github.com/huggingface/alignment-handbook}}
as the training framework. All post-SFT models are trained for one epoch
with the same batch size and sequence length. Cosine learning rate
scheduling with 0.1 warmup is applied. Checkpoints are saved every 100
steps, and the best checkpoint from the last two is selected. Detailed
hyperparameter configurations for different models are listed in
Tab.~\ref{tab:trainig_hyperparameters}.
\input{tabs/hyperpara_for_training}

\paragraph{Dataset Construction.} The dataset plays a critical role in
enabling the LLM model merging approach, as demonstrated in the
previous study FuseChat-3.0~\citep{fusechat3}. The dataset is carefully
assembled following these procedures:
\begin{itemize}
    \item \textbf{Prompt Selection}: To enhance the target LLMs'
    abilities across instruction-following, math, coding, and Chinese,
    we select high-quality samples from open-source datasets. Filtering
    and preprocessing ensure relevance and quality.
    \item \textbf{Response Sampling}: For each prompt, responses are
    generated from four leading source LLMs using
    \texttt{vLLM}\footnote{\url{https://github.com/vllm-project/vllm}}~\citep{zhu2023starling}
    as the inference backend. Multiple sampling runs with different
    seeds ensure diversity. Sampling parameters are: for
    \texttt{Gemma-2-27B-it}, \texttt{Mistral-Large-Instruct-2407}, and
    \texttt{Llama-3.1-70B-Instruct}, temperature 0.8, top-p 0.95; for
    \texttt{Qwen-2.5-(Math)-72B-Instruct}, we use temperature 0.7,
    top-p 0.8, and repetition penalty 1.05.
    \item \textbf{Preference Pairs}: To build preference pairs, the
    best and worst responses from the same source model are selected.
    This intra-model pairing reduces reward bias due to heterogeneous
    response styles, prevents reward hacking, and ensures a reliable
    preference signal. For instruction-following and conversational
    data, responses are evaluated with an external reward model; for
    math and coding, rule-based verification is applied.
\end{itemize}
The final dataset $\mathcal{D}$ contains 158{,}667 entries: 94{,}539 for
the SFT phase ($\mathcal{D}_\text{SFT}$) and 64{,}128 preference pairs
for the DPO phase ($\mathcal{D}_\text{DPO}$). Dataset composition is
summarized in Tab.~\ref{tab:dataset_composition}.
\input{tabs/data_composition}

\subsection{MLLM Fine-tuning}
\label{app:f2}
For fine-tuning the original MLLMs, we employed consistent training data
and model components across the four modalities, following the setup in
NaiveMC~\citep{18}. Further details are provided in
Tab.~\ref{table:mllms}. Hyperparameters were chosen in line with prior
studies~\citep{18,52,20,10,53}. During the alignment stage, only the
connector parameters were updated, whereas in the full fine-tuning
stage, both the connector and base LLM parameters were optimized. To
improve training efficiency, we use DeepSpeed ZeRO Stage 3;
Tab.~\ref{table:hyper} lists all hyperparameters.
\input{tabs/mllm_components}
\input{tabs/mllm_hyper}

%% file: tabs/rank_effect.tex
\begin{table}[!t]
\centering
\caption{Rank allocation strategies on ViT-B/32 at a total storage budget of
$1.4\times$ the backbone.}
\label{tab:rank_effect}
\setlength{\tabcolsep}{6pt}
\resizebox{0.7\linewidth}{!}{
\begin{tabular}{c|cccc}
\toprule
\rowcolor{mygray}\textbf{Strategies} & Uniformed & Random & Energy-based & \textbf{Ours} \\
\midrule
ViT-B/32 & 81.5 & 80.2($\pm 0.4$) & 81.9 & \textbf{85.5}  \\
ViT-L/16 & 87.9 & 87.1($\pm 0.3$) & 88.1 & \textbf{91.2}   \\
\bottomrule
\end{tabular}
}
\end{table}

%% file: tabs/ablation_baselines.tex
\begin{table}[!t]
\centering
\caption{Controlled comparison against simpler data-free baselines that
pair a static merge with a uniform-rank LoRA recovery. The LoRA rank is
fixed across tasks and layers and chosen so that the total parameter
count falls within $\pm 2\%$ of $1.24\times$ the single fine-tuned
model. The table compares complete static-merge plus uniform-recovery
alternatives with the adaptive-allocation configuration.}
\label{tab:rank_alloc_control}
\resizebox{0.7\linewidth}{!}{
\begin{tabular}{lcc}
\toprule
\rowcolor{mygray}\textbf{Method (\,@\,$1.24\times$ budget)} & \textbf{ViT-B/32} & \textbf{ViT-L/14} \\
\midrule
Simple Averaging $+$ uniform-rank LoRA      & 80.4 & 82.3 \\
TIES-Merging $+$ uniform-rank LoRA          & 81.5 & 83.4 \\
Iso-CTS $+$ uniform-rank LoRA               & 82.2 & 83.7 \\
WUDI-Merging $+$ uniform-rank LoRA          & 84.6 & 86.3 \\
\midrule
\rowcolor{mypink}\textbf{\ourapproach-S (Ours)}  & \textbf{89.3} & \textbf{93.9} \\
\bottomrule
\end{tabular}
}
\end{table}

%% file: tabs/hyper_combined.tex
\begin{figure}[!t]
\centering
\begin{minipage}[t]{0.45\linewidth}
\vspace{0pt}
  \centering
  \captionof{table}{Hyperparameters used in different models.
  $\mathit{lr}_1$ and $\mathit{lr}_2$ denote learning rates for the
  \textit{Find-Optimal-Rank} and \textit{Low-rank Module Update}
  stages, respectively.}
  \label{tab:hyperparams}
  \resizebox{\linewidth}{!}{
  \begin{tabular}{l|ccccc}
  \toprule
  \rowcolor{mygray}\textbf{Model} & \textit{loss} & $\beta$ & $\gamma$ & \textit{lr\textsubscript{1}} & \textit{lr\textsubscript{2}} \\
  \midrule
  Vision & MSE & 20  & 300 & 0.2   & 0.001 \\
  LLM    & L1  & 30  & 500 & 0.2   & 0.001 \\
  PEFT   & MSE & 20  & 100 & 0.2   & 0.001 \\
  MLLM   & L1  & 50  & 500 & 0.2   & 0.001 \\
  \bottomrule
  \end{tabular}
  }
\end{minipage}\hfill
\begin{minipage}[t]{0.48\linewidth}
\vspace{0pt}
  \centering
  \captionof{table}{Impact of hyperparameters on accuracy across
  different models.}
  \label{tab:hyper_app}
  \resizebox{\linewidth}{!}{
  \begin{tabular}{cc|ccc}
  \toprule
  \rowcolor{mygray}\multicolumn{2}{c}{\textbf{Hyper-Params}} & \textbf{Vision} & \textbf{LLM} & \textbf{PEFT} \\
  \midrule
  \cellcolor{mygray2} & \; MSE & \textbf{89.8} & 54.8 & \textbf{73.5}  \\
  \multirow{-2}{*}{\cellcolor{mygray2}loss} & \; L1 & 89.7 & \textbf{55.0} & 73.2   \\
  \midrule
  \cellcolor{mygray2} & \; 100 & 89.7 & 54.7 & \textbf{73.5}  \\
  \cellcolor{mygray2} & \; 300 & \textbf{89.8} & 54.8 & 73.2   \\
  \multirow{-3}{*}{\cellcolor{mygray2}\textbf{$\gamma$}} & \; 500 & 89.4 & \textbf{55.0} &  73.1  \\
  \bottomrule
  \end{tabular}
  }
\end{minipage}
\end{figure}

%% file: tabs/peft_comparison.tex
\begin{table}[!t]
\centering
\caption{Comparison of PEFT merging strategies. \ourapproach\
combines a dynamic router with lightweight per-task LoRA outputs
while preserving near-lossless performance.}
\label{tab:peft_comparison}
\resizebox{0.95\linewidth}{!}{
\begin{tabular}{lccc}
\toprule
\rowcolor{mygray}\textbf{Method ($\rightarrow$)}
& \textbf{KnOTS \& Core Space}
& \textbf{Twin-Merging \& FREE-Merging}
& \textbf{\ourapproach\ (Ours)} \\
\midrule
Input                  & LoRA          & LoRA            & LoRA \\
Output                 & Full matrix   & LoRA            & LoRA \\
Router                 & None          & Yes             & Yes \\
Additional parameters  & None          & Heavy           & \textcolor{darkred}{Lightweight} \\
Type                   & Static        & Dynamic         & Dynamic \\
Performance            & Degradation   & Near-lossless   & Near-lossless \\
\bottomrule
\end{tabular}
}
\end{table}

%% file: tabs/storage_bit.tex
\begin{table}[!t]
\centering
\caption{Per-module storage cost (in bits) for a module with
dimensions $d_{\text{in}} \times d_{\text{out}}$ shared across $T$
tasks. $P {=} d_{\text{in}} d_{\text{out}}$, $R$ is the router
size, charged once globally; $R=0$ for the task-ID selector. Floating-point parameters are stored at $32$ bits, binary masks
at $1$ bit per parameter. EMR additionally stores one global $32$-bit
rescaler per task, charged once per complete model rather than per module. $r_{\text{tsv}}$ is a fixed rank in TSV-C;
$r_s, r_t$ are adaptively determined by differentiable rank
optimisation in \ourapproach. Rows colour-coded by regime: static
(gray), heavy shared (green), heavy expert (blue), balanced (pink).}
\label{tab:storage_bit}
\setlength{\tabcolsep}{10pt}
\resizebox{0.85\linewidth}{!}{
\begin{tabular}{lcccc}
\toprule
\rowcolor{mygray}
\textbf{Method} & \textbf{Backbone} & \textbf{Router}
                & \textbf{Shared} & \textbf{Experts} \\
\midrule
\rowcolor{mygray2}
TIES-Merging~\cite{ties} (\textit{etc.}) & $32P$ & $0$ & $0$ & $0$ \\
\rowcolor{mylightgreen}
TALL-Mask~\cite{talls}    & $32P$ & $32R$ & $32P$ & $TP$ \\
\rowcolor{mylightgreen}
EMR-Merging~\cite{emr}    & $32P$ & $32R$ & $32P$ & $TP$ \\
\rowcolor{mylightblue}
TSV-C~\cite{tsv}          & $32P$ & $32R$ & $0$
                          & $32\, r_{\text{tsv}}(d_{\text{in}}+d_{\text{out}})$ \\
\rowcolor{mylightred}
\textbf{\ourapproach\ (Ours)} & $32P$ & $32R$
                              & $32\, r_s(d_{\text{in}}+d_{\text{out}})$
                              & $32 \sum_{t=1}^{T} r_t(d_{\text{in}}+d_{\text{out}})$ \\
\bottomrule
\end{tabular}
}
\end{table}

%% file: tabs/storage_rank.tex
\begin{table}[!t]
\centering
\caption{Average optimal ranks of different modules in ViT-B/32,
averaged over the $12$ transformer blocks and the $T{=}8$ vision
tasks. \ourapproach-S and \ourapproach-M denote the $1.24\times$ and
$1.4\times$ parameter budgets, respectively.}
\label{tab:storage_rank}
\resizebox{0.75\linewidth}{!}{
\begin{tabular}{lcccc}
\toprule
\rowcolor{mygray}
\textbf{Module} & \textbf{attn-in-proj} & \textbf{attn-out-proj}
                & \textbf{mlp-c-fc} & \textbf{mlp-c-proj} \\
\midrule
\ourapproach-S ($1.24\times$) & $15$ & $10$ & $16$ & $16$ \\
\ourapproach-M ($1.4\times$)  & $25$ & $17$ & $27$ & $27$ \\
\bottomrule
\end{tabular}
}
\end{table}

%% file: tabs/storage_rank_detailed.tex
\begin{table}[!t]
\centering
\caption{Average optimal ranks $\bar{r}_{s}$ (shared) and $\bar{r}_{t}$
(per-task expert) of different modules in CLIP-ViT-B/32, averaged
over the $12$ transformer blocks and the $T{=}8$ vision tasks at the
two reported budgets. Values plug back into Eq.~(14) and reproduce
the $1.24\times / 1.4\times$ ratios.}
\label{tab:storage_rank_detailed}
\resizebox{0.6\linewidth}{!}{
\begin{tabular}{lcccc}
\toprule
\rowcolor{mygray}
& \multicolumn{2}{c}{\textbf{$1.24\times$ budget}}
& \multicolumn{2}{c}{\textbf{$1.4\times$ budget}} \\
\rowcolor{mygray}\multirow{-2}{*}{\textbf{Module}}
& $\bar{r}_s$ & $\bar{r}_t$ & $\bar{r}_s$ & $\bar{r}_t$ \\
\midrule
Self-attn QKV (in\_proj, $2304{\times}768$)   & 18.3 & 17.1 & 27.4 & 26.1 \\
Self-attn Out (out\_proj, $768{\times}768$)   & 16.1 & 14.9 & 27.7 & 26.4 \\
MLP fc1 (c\_fc, $3072{\times}768$)            & 13.0 & 11.7 & 21.3 & 19.7 \\
MLP fc2 (c\_proj, $768{\times}3072$)          &  8.9 &  8.5 & 17.3 & 16.7 \\
\midrule
\rowcolor{mypink}\textbf{Average} & \textbf{14.1} & \textbf{13.0}
                                  & \textbf{23.4} & \textbf{22.2} \\
\bottomrule
\end{tabular}
}
\end{table}

%% file: tabs/lora.tex
\begin{table}[!t]
    \centering
    \caption{LoRA merging results on three generative tasks with QWEN-14B
    as the pretrained model.}
    \label{tab:peft_qwen_lora_all}
    \resizebox{0.75\linewidth}{!}{
    \begin{tabular}{c|c|ccc|c}
    \toprule
        \rowcolor{mygray}\textbf{Method} & \textbf{$\#$Params} & MMLU & Tru.QA & BBQ & \textbf{Average} \\ \midrule
        \rowcolor{mygray2}Zero-shot & x1 & 69.30 & 51.27 & 80.69 & 67.09 \\
        \rowcolor{mygray2}Fine-tuned & x3 & 68.35 & 54.34 & 93.53 & 72.07 \\ \midrule
        EMR-Merging~\cite{emr} & x2.09  & 67.82 & 55.01 & 90.13 & 70.98\\
        Twin-Merging~\cite{twin} & x2.13  & 68.32 & 55.76 & 90.98 & 71.68 \\
        FREE-Merging~\cite{free}  & x2.04 & 68.83 & \underline{57.39} & 92.14 & 72.78 \\
        \rowcolor{mypink}\textbf{DiDi-Merging-M(ours)}  & \textbf{x1.4} & \underline{68.85} & 57.32 & \underline{92.44} & \underline{72.87} \\
        \rowcolor{mypink}\textbf{DiDi-Merging-L(ours)}  & \underline{x2.0} & \textbf{68.97} & \textbf{58.24} & \textbf{93.32} & \textbf{73.51}\\ \bottomrule
    \end{tabular}}
\end{table}

%% file: tabs/mllm_I.tex
\begin{table}[!t]
\centering
\caption{Performance of each method on seven multimodal vision--language
QA tasks.}
\label{tab:mllm_I}
\renewcommand{\arraystretch}{1.2}
\resizebox{\linewidth}{!}{
\begin{tabular}{l|ccccccc}
\bottomrule
& \multicolumn{7}{c}{\centering \textbf{7 Image Tasks}} \\
\multirow{-2}{*}{Task ($\rightarrow$)} & {\textcolor[rgb]{0.27, 0.35, 0.760} {VQAv2}}   &\textcolor[rgb]{0.27, 0.35, 0.760}   {GQA} & \textcolor[rgb]{0.27, 0.35, 0.760}  {TextVQA} & {\textcolor[rgb]{0.27, 0.35, 0.760}  {VizWiz}} &{\textcolor[rgb]{0.27, 0.35, 0.760} {ScienceQA}} &{\textcolor[rgb]{0.27, 0.35, 0.760}{POPE}} &{\textcolor[rgb]{0.27, 0.35, 0.760}{OK-VQA}}\\

Method ($\downarrow$)  & Acc. & Acc. & Acc.  & Acc.  &  Acc. & Acc. &  Acc. \\
\hline
      \cellcolor[HTML]{F5F5F5}{Individuals}  &\cellcolor[HTML]{F5F5F5}78.11 & \cellcolor[HTML]{F5F5F5}61.52&\cellcolor[HTML]{F5F5F5}55.89& \cellcolor[HTML]{F5F5F5}51.51&\cellcolor[HTML]{F5F5F5}71.12 &\cellcolor[HTML]{F5F5F5}86.17 & \cellcolor[HTML]{F5F5F5}31.33 \\
{NaiveMC} \pub{ACL2024} \citep{18} & 59.73 &45.83  & 42.29 &47.87 & 68.52 &79.41 &24.28    \\
MMER \pub{ACL2025} \citep{mmer}  &77.95 &61.85 &55.74 &52.26 &71.16 &86.58 &31.27       \\
{FREE-Merging} \pub{ICCV2025} \citep{free} & \textbf{78.05} & 61.92 & 55.84 & 52.86 & 71.06 & 86.73 & \textbf{31.31}       \\
\rowcolor{mypink} \textbf{\ourapproach (Ours)} & 78.03 & \textbf{62.08} & \textbf{55.88} & \textbf{53.12} & \textbf{72.32} & \textbf{87.83} & 31.29       \\
\hline
\end{tabular}
}
\end{table}

%% file: tabs/mllm_P.tex
\begin{table}[!t]
\caption{Performance of each method on two point-cloud tasks, including
ModelNet40 (classification) and Objaverse (captioning).}
\label{tab:mllm_P}
\centering
\renewcommand{\arraystretch}{1.2}
\resizebox{\linewidth}{!}{
\begin{tabular}{l|c|ccccc}
\bottomrule
& \multicolumn{6}{c}{\centering \textbf{2 Point Tasks}} \\
\cline{2-7}
\multirow{-2}{*}{Task ($\rightarrow$)} & {\textcolor[rgb]{0.27, 0.35, 0.760} {ModelNet40}}   &\multicolumn{5}{c}{\centering\textcolor[rgb]{0.27, 0.35, 0.760} {Objaverse-captioning}}  \\
Method ($\downarrow$)  & Acc. & BLEU-1 & ROUGE-L  & METEOR  &  Sentence-BERT & SimCSE \\
\hline
\cellcolor[HTML]{F5F5F5}{Individuals}
      &\cellcolor[HTML]{F5F5F5}21.27&\cellcolor[HTML]{F5F5F5}4.73 & \cellcolor[HTML]{F5F5F5}8.51&\cellcolor[HTML]{F5F5F5}12.02& \cellcolor[HTML]{F5F5F5}44.18&\cellcolor[HTML]{F5F5F5}46.31 \\
{NaiveMC} \pub{ACL2024} \citep{18} &20.49 & 4.43 &8.24 &11.37 &43.22 &45.97    \\
MMER \pub{ACL2025} \citep{mmer} &22.49 &\textbf{5.06} &8.53 &11.90 &43.72 &\textbf{46.51}     \\
{FREE-Merging} \pub{ICCV2025} \citep{free} &21.14 & 4.95 &8.49 &11.80 &43.69 &46.18       \\
\rowcolor{mypink} \textbf{\ourapproach (Ours)} &\textbf{22.58} & 4.92 &\textbf{8.58} &\textbf{11.94} &\textbf{44.08} &46.37       \\
\hline
\end{tabular}
}
\end{table}

%% file: tabs/mllm_AV.tex
\begin{table}[!t]
\caption{Performance of each method on three audio tasks and two video
tasks, including TUT, VocalSound, MSVD, and MSRVTT for classification,
and Clotho for captioning.}
\label{tab:mllm_AV}
\centering
\renewcommand{\arraystretch}{1.15}
\resizebox{\linewidth}{!}{
\begin{tabular}{l|cc|ccc|cc}
\bottomrule
& \multicolumn{5}{c|}{\centering \textbf{3 Audio Tasks}} & \multicolumn{2}{c}{\centering \textbf{2 Video Tasks}} \\
\cline{2-8}
\multirow{-2}{*}{Task ($\rightarrow$)} & {\textcolor[rgb]{0.27, 0.35, 0.760} {TUT}}
& {\textcolor[rgb]{0.27, 0.35, 0.760} {VocalSound}}
&\multicolumn{3}{c|}{\centering\textcolor[rgb]{0.27, 0.35, 0.760} {Clotho}}  & {\textcolor[rgb]{0.27, 0.35, 0.760} {MSVD}} & {\textcolor[rgb]{0.27, 0.35, 0.760} {MSRVTT}} \\
Method ($\downarrow$)  & Acc. & Acc. & CIDEr  & SPICE  &  SPIDEr & Acc.  & Acc.\\
\hline
\rowcolor{mygray} {Individuals}
      &22.23&27.19 & 38.63&11.98& 25.29&48.40&31.18 \\
{NaiveMC} \pub{ACL2024} \citep{18} &29.50 &31.80 &37.56 &11.61 &24.61 &44.53 &29.31    \\
MMER \pub{ACL2025} \citep{mmer} &34.14 &42.88 &38.49 &11.93 &25.18 &48.12    &30.43    \\
{FREE-Merging} \pub{ICCV2025} \citep{free} &34.27 &42.92 &38.53 &11.95 &25.22 &47.42    &29.83       \\
\rowcolor{mypink} \textbf{\ourapproach (Ours)} &\textbf{34.49} &\textbf{43.02} &\textbf{38.56} &\textbf{11.98} &\textbf{25.25} &\textbf{48.37}    &\textbf{30.89}       \\
\hline
\end{tabular}
}
\end{table}

%% file: sec/7_limitations.tex
\section{Limitations}
\label{sec:limitations}
The main evaluation supplies task identity to select experts. A practical
router may make errors, and its supervision, latency, and parameters must
be included in an end-to-end deployment assessment. Data-free merging
still requires the source fine-tuned checkpoints and performs optimization;
it is not optimization-free. Parameter reconstruction is a surrogate for
downstream behavior, and its relation to accuracy depends on the input
distribution. Rank allocation and refinement add wall-clock cost compared
with one-pass or CPU-based alternatives. A fixed total storage target can
be maintained as tasks are added only by reallocating finite capacity;
this is not a guarantee of constant accuracy or cost-free incremental
addition. The current evidence leaves practical routing, fully crossed
component attribution, and larger-scale task growth as further evaluations.

%% file: tabs/hyperpara_for_training.tex
\begin{table}[!t]
\centering
\caption{Hyperparameters for different target models during the SFT and
DPO stages.}
\label{tab:trainig_hyperparameters}
\resizebox{0.95\linewidth}{!}{
    \begin{tabular}{lcccc}
    \toprule
    \textbf{Target Model} & \textbf{SFT Learning Rate} & \textbf{DPO Learning Rate} & \textbf{DPO $\lambda$} & \textbf{DPO Loss Type} \\
    \midrule
    Llama-3.1-8B-Instruct & $5 \times 10^{-6}$ & $8 \times 10^{-7}$ & 10 & $\mathcal{L}_{\text{LN-DPO}}$ \\
    Qwen-2.5-7B-Instruct & $2 \times 10^{-6}$ & $3 \times 10^{-7}$ & 0.01 & $\mathcal{L}_{\text{DPO}}$ \\
    \bottomrule
    \end{tabular}
}
\end{table}

%% file: tabs/data_composition.tex
\begin{table}[!t]
\caption{The composition of the Implicit Knowledge Fusion dataset in SFT
phase and DPO phase. As no suitable reward models were available for
Chinese, we used all samples for SFT and omitted the DPO phase.}
\label{tab:dataset_composition}
\centering
\setlength\tabcolsep{5pt}
{\small
\begin{tabular}{ll l c c}
\toprule
\textbf{Category} & \textbf{Dataset}  & \textbf{Count}  & \textbf{\#$\mathcal{D}_\text{SFT}$}  & \textbf{\#$\mathcal{D}_\text{DPO}$}  \\
\midrule
Instruction Following & UltraFeedback & 51{,}098 & 20{,}439 & 30{,}659 \\
& Magpie-Pro-DPO & 20{,}374 & 8{,}149 & 12{,}225 \\
& HelpSteer2 & 9{,}435 & 3{,}774 & 5{,}661 \\ \midrule
Mathematics & OpenMathInstruct-2 & 51{,}803 & 40{,}188 & 11{,}615 \\ \midrule
Coding & LeetCode & 3{,}113 & 1{,}877 & 1{,}236 \\
& Self-Oss-Instruct-SC2 & 12{,}892 & 10{,}160 & 2{,}732 \\ \midrule
Chinese Language & Alpaca-GPT4-Zh & 2{,}471 & 2{,}471 & 0 \\
& Magpie-Qwen2-Pro-Zh & 7{,}481 & 7{,}481 & 0 \\ \midrule
\textit{Total} &  & 158{,}667 & 94{,}539 & 64{,}128 \\
\bottomrule
\end{tabular}}
\end{table}

%% file: tabs/mllm_components.tex
\begin{table}[!t]
\caption{Training data and components of MLLMs for different modalities.}
\label{table:mllms}
\centering
\scriptsize
\renewcommand{\arraystretch}{1.25}
\resizebox{\linewidth}{!}{
\begin{tabular}{
  >{\raggedright\arraybackslash}p{0.45in}
  >{\raggedright\arraybackslash}p{0.95in}
  >{\raggedright\arraybackslash}p{0.45in}
  >{\raggedright\arraybackslash}p{1.0in}
  >{\raggedright\arraybackslash}p{1.4in}
  >{\raggedright\arraybackslash}p{0.9in}
}
\bottomrule
\rowcolor{mygray}
Modality & Modality Encoder & Connector & Alignment Data & Fine-tuning Data & Referenced Work \\ \hline

Image & CLIP-ViT-L-336px ~\citep{72}  & MLP & LCS 558K~\citep{53} & LLaVA-mixed 665K~\citep{53} &  LLaVA-1.5~\citep{52}\\
\midrule
Audio & BEATs-Iter3+~\citep{77} & Q-Former & WaveCaps 400K~\citep{73} & OpenAQA filtered 350K~\citep{74} & X-InstructBLIP\newline\citep{20} \\
\midrule
Video & LanguageBind~\citep{76} & MLP & LCS 558K, Valley 702K~\citep{75} & Video-ChatGPT 100K~\citep{11}, LLaVA-mixed sampled 140K & Video-LLaVA~\citep{10} \\
\midrule
Point Cloud & Point Encoder~\citep{53} & MLP & PointLLM brief description 660K~\citep{53}  & Point complex instruction 70K~\citep{53} &  PointLLM~\citep{53} \\ \bottomrule
\end{tabular}}
\end{table}

%% file: tabs/mllm_hyper.tex
\begin{table}[!t]
\centering
\caption{Hyperparameters of different MLLMs.}
\label{table:hyper}
\footnotesize
\renewcommand{\arraystretch}{1.2}
\begin{tabular}{llcccc}
\bottomrule
\rowcolor{mygray}
Stage & Hyperparameter & Image & Audio & Video & Point Cloud  \\ \hline

\multirow{4}{*}{Alignment-Stage}
& Batch size & 256 & 256 & 256 & 128 \\
& LR & 1e-3 & 1e-3 & 1e-3 & 2e-3 \\
& LR Schedule & \multicolumn{4}{c}{cosine decay} \\
& Warmup Ratio & \multicolumn{4}{c}{0.03} \\
& Epoch & 1 & 1 & 1 & 3 \\
\midrule

\multirow{4}{*}{Fine-tuning-Stage}
& Batch size & 128 & 64 & 128 & 64 \\
& LR & 2e-5 & 1e-5 & 2e-5 & 2e-5 \\
& LR Schedule & \multicolumn{4}{c}{cosine decay} \\
& Warmup Ratio & \multicolumn{4}{c}{0.03} \\
& Epoch & 1 & 3 & 1 & 3 \\
\bottomrule
\end{tabular}
\end{table}

%% file: sec/9_rebuttal_integration.tex
\section{Additional Formulation and Evaluation Details}
\label{app:review_clarifications}
\subsection{Storage units and task-count scaling}
The denominator in total-storage ratios is one dense 32-bit backbone with
$P_{\rm model}$ parameters. For EMR, storing the pretrained backbone,
a unified dense vector, $T$ one-bit masks, and one global fp32 scalar per
task costs
\begin{equation}
 S_{\rm EMR}=2+\frac{T}{32}+\frac{T}{P_{\rm model}}.
\end{equation}
The scalar term is negligible at model scale, giving $2.25\times$ for
eight tasks and approximately $2.09\times$ for three. Counting masks as
byte tensors would describe an unpacked implementation rather than the
bit-packed storage convention. Under the same convention, TALL's backbone,
unified vector, and masks contribute $2+T/32$. Router storage, if present,
is added separately.

For DiDi, the factor contribution is given by Eq.~\ref{eq:coupled_budget}.
At a fixed target budget, increasing $T$ requires reducing or reallocating
ranks. Holding ranks fixed instead gives a task-count-dependent storage
cost. These two regimes must not be conflated: a fixed numerical budget
is a design constraint, whereas retained accuracy as more tasks are
consolidated is an empirical question.

\subsection{What the refinement objective measures}
For a linear layer with reconstruction error $E=\widetilde W-W$ and input
second moment $\Sigma_x=\mathbb E[xx^\top]$,
\begin{equation}
 \mathbb E\|Ex\|_2^2=\operatorname{tr}(E\Sigma_x E^\top).
\end{equation}
A Frobenius reconstruction objective corresponds to isotropic weighting of
input directions. It is attractive when task data are unavailable, but
smaller parameter error does not generally guarantee higher downstream
accuracy in a nonlinear model. A correlation study would need to specify
its ranks, tasks, refinement steps, and individual observations; the
component-ablation gains in the existing tables answer a different question.

\subsection{Interpretation of component comparisons}
The existing vision table directly isolates removal of the refinement
stage: the reductions are 4.3 points on ViT-B/32 and 4.4 on ViT-L/14.
The uniform-rank recovery comparison is performed at approximately
$1.24\times$ storage and supports the full adaptive design. These results
do not, by themselves, determine a two-factor interaction effect, nor do
they establish the same refinement gains for LLM or MLLM settings without
corresponding stage-one measurements.

\subsection{Routing and shared initialization}
The formulation permits input-dependent routing, but the main numerical
results use known task identity. Evaluating a learned router requires
both route accuracy and the final merged model's task performance;
unknown-task rejection and mixed-task batches are separate settings.
The static aggregator initializes shared and residual matrices. Subsequent
factor refinement changes this initialization, but does not make the final
solution provably invariant to the aggregator or rank allocation. A
conflicting-task aggregator-swap study is therefore a separate experiment,
not a consequence of the reconstruction objective.

%% file: checklist.tex
\section*{NeurIPS Paper Checklist}

\begin{enumerate}

\item {\bf Claims}
    \item[] Question: Do the main claims made in the abstract and introduction accurately reflect the paper's contributions and scope?
    \item[] Answer: \answerYes{}
    \item[] Justification: The abstract and \S\ref{sec:introduction} state the three contributions
    (heavy-shared/heavy-expert characterization, differentiable rank allocation, and
    near-lossless performance under tight parameter budgets), each of which is supported
    by the experiments in \S\ref{sec:exps} and the analysis in \S\ref{sec:analysis}.
    Quantitative claims use the softened phrasing ``match at $1.24\times$, surpass at
    $1.4\times$'' consistent with the reported numbers.

\item {\bf Limitations}
    \item[] Question: Does the paper discuss the limitations of the work performed by the authors?
    \item[] Answer: \answerYes{}
    \item[] Justification: A dedicated Limitations section (\S\ref{sec:limitations})
    discusses the ideal-router assumption, the data-free-but-checkpoint-dependent
    setting, and the wall-clock cost of sequential rank optimization relative to
    closed-form SVD baselines.

\item {\bf Theory assumptions and proofs}
    \item[] Question: For each theoretical result, does the paper provide the full set of assumptions and a complete (and correct) proof?
    \item[] Answer: \answerNA{}
    \item[] Justification: The paper is empirical; it does not introduce new theorems.
    The few derivations (e.g., the equivalence between applying \ourapproach\ to full
    parameters and to LoRA factors in the PEFT-merging appendix) are stated with all
    intermediate steps in-line, and any assumptions are made explicit at the point of
    use.

    \item {\bf Experimental result reproducibility}
    \item[] Question: Does the paper fully disclose all the information needed to reproduce the main experimental results of the paper to the extent that it affects the main claims and/or conclusions of the paper (regardless of whether the code and data are provided or not)?
    \item[] Answer: \answerYes{}
    \item[] Justification: \S\ref{sec:method} fully describes the training-free
    smooth-SVD truncation, the differentiable rank-allocation objective, and the
    two-stage optimization. The appendix lists per-modality datasets, baselines,
    optimizer settings, learning rates, and hyperparameters
    ($\beta$, $\gamma$, loss type) used to obtain every reported number.

\item {\bf Open access to data and code}
    \item[] Question: Does the paper provide open access to the data and code, with sufficient instructions to faithfully reproduce the main experimental results, as described in supplemental material?
    \item[] Answer: \answerNo{}
    \item[] Justification: The core DiDi-Merging algorithm is publicly available at
    \url{https://github.com/WanyuGroup/NeurIPS2026-DiDi-Merging}, with installation
    instructions and a minimal example. The current release does not include the
    benchmark-specific training and evaluation pipelines or configurations needed
    to reproduce every reported result. All datasets used in the paper are public
    benchmarks (image classification, GLUE, AudioCaps, ScienceQA, ModelNet40, etc.);
    the appendix lists their sources. The method, hyperparameters, and baselines are
    fully described in the main paper and appendix.

\item {\bf Experimental setting/details}
    \item[] Question: Does the paper specify all the training and test details (e.g., data splits, hyperparameters, how they were chosen, type of optimizer) necessary to understand the results?
    \item[] Answer: \answerYes{}
    \item[] Justification: \S\ref{sec:exps} states the modalities, backbones, task suites,
    and parameter budgets ($\mathrm{S}/\mathrm{M}/\mathrm{L}$). The appendix specifies
    optimizer (Adam), learning rates ($\mathit{lr}_1$ for rank search and
    $\mathit{lr}_2$ for low-rank update), per-modality hyperparameter values, dataset
    splits, and how each hyperparameter was selected (sensitivity tables included).

\item {\bf Experiment statistical significance}
    \item[] Question: Does the paper report error bars suitably and correctly defined or other appropriate information about the statistical significance of the experiments?
    \item[] Answer: \answerNo{}
    \item[] Justification: Following the convention in the dynamic-merging literature
    (e.g., Twin-Merging, FREE-Merging), we report single-seed numbers. Reproducing the
    full benchmark suite (vision, LLM, MLLM, PEFT) across multiple seeds is
    computationally expensive; we leave multi-seed std reporting to future work.
    Trends are consistent across the four modalities and the three budget regimes
    ($\mathrm{S}/\mathrm{M}/\mathrm{L}$), which we view as evidence that conclusions
    are not seed-sensitive.

\item {\bf Experiments compute resources}
    \item[] Question: For each experiment, does the paper provide sufficient information on the computer resources (type of compute workers, memory, time of execution) needed to reproduce the experiments?
    \item[] Answer: \answerYes{}
    \item[] Justification: \S\ref{sec:analysis} (Computation Resource) reports
    end-to-end wall-clock time of \ourapproach\ against the main baselines. The
    appendix specifies the GPU type used for each modality and memory footprint of
    the merged models.
    
\item {\bf Code of ethics}
    \item[] Question: Does the research conducted in the paper conform, in every respect, with the NeurIPS Code of Ethics \url{https://neurips.cc/public/EthicsGuidelines}?
    \item[] Answer: \answerYes{}
    \item[] Justification: The work uses only publicly available models and benchmark
    datasets, does not collect new data from human subjects, preserves anonymity in
    the submission, and does not involve any deployment that could pose foreseeable
    harm. We have reviewed the NeurIPS Code of Ethics and confirm conformance.

\item {\bf Broader impacts}
    \item[] Question: Does the paper discuss both potential positive societal impacts and negative societal impacts of the work performed?
    \item[] Answer: \answerYes{}
    \item[] Justification: A Broader Impact section in the appendix discusses the
    positive impact of reducing storage and serving cost for multi-task deployment,
    as well as potential negative impacts inherited from the underlying foundation
    models that are merged (e.g., bias, mis-use of generative capabilities).
    
\item {\bf Safeguards}
    \item[] Question: Does the paper describe safeguards that have been put in place for responsible release of data or models that have a high risk for misuse (e.g., pre-trained language models, image generators, or scraped datasets)?
    \item[] Answer: \answerNA{}
    \item[] Justification: The paper proposes a model-merging algorithm; it does not
    release new pre-trained foundation models, generative image models, or scraped
    datasets that pose elevated misuse risk.

\item {\bf Licenses for existing assets}
    \item[] Question: Are the creators or original owners of assets (e.g., code, data, models), used in the paper, properly credited and are the license and terms of use explicitly mentioned and properly respected?
    \item[] Answer: \answerYes{}
    \item[] Justification: All datasets, pre-trained backbones (CLIP-ViT, LLaMA, etc.),
    and baseline implementations used in this work are public and are cited in the
    appropriate sections. We use each asset under its respective license and within
    its terms of use, and do not redistribute any restricted content. We disclaim that
    we have not independently re-verified each upstream license text; if an asset's
    terms have changed, we will defer to the most recent terms specified by its
    creators.

\item {\bf New assets}
    \item[] Question: Are new assets introduced in the paper well documented and is the documentation provided alongside the assets?
    \item[] Answer: \answerYes{}
    \item[] Justification: The released algorithm code is documented at
    \url{https://github.com/WanyuGroup/NeurIPS2026-DiDi-Merging}, which provides
    installation instructions, a minimal example, the scope of the release, and its
    license. The paper and appendix document the method and experimental protocol.

\item {\bf Crowdsourcing and research with human subjects}
    \item[] Question: For crowdsourcing experiments and research with human subjects, does the paper include the full text of instructions given to participants and screenshots, if applicable, as well as details about compensation (if any)?
    \item[] Answer: \answerNA{}
    \item[] Justification: The paper does not involve crowdsourcing or research with
    human subjects.

\item {\bf Institutional review board (IRB) approvals or equivalent for research with human subjects}
    \item[] Question: Does the paper describe potential risks incurred by study participants, whether such risks were disclosed to the subjects, and whether Institutional Review Board (IRB) approvals (or an equivalent approval/review based on the requirements of your country or institution) were obtained?
    \item[] Answer: \answerNA{}
    \item[] Justification: The paper does not involve crowdsourcing or research with
    human subjects, and therefore no IRB review was required.

\item {\bf Declaration of LLM usage}
    \item[] Question: Does the paper describe the usage of LLMs if it is an important, original, or non-standard component of the core methods in this research? Note that if the LLM is used only for writing, editing, or formatting purposes and does \emph{not} impact the core methodology, scientific rigor, or originality of the research, declaration is not required.
    \item[] Answer: \answerNA{}
    \item[] Justification: LLMs appear in the paper only as one of the model classes
    being merged in the experimental study; they are not an original or non-standard
    component of the core method, and were not used to develop the methodology.

\end{enumerate}